\PassOptionsToPackage{table}{xcolor}
\documentclass[10pt,twocolumn,letterpaper]{article}

\usepackage{cvpr}              

\usepackage[dvipsnames]{xcolor}

\usepackage{amsmath}
\usepackage{amsfonts}
\usepackage{bm}

\def\eqref#1{equation~\ref{#1}}

\def\1{\bm{1}}

\def\vx{{\bm{x}}}
\def\vy{{\bm{y}}}
\def\vz{{\bm{z}}}

\DeclareMathAlphabet{\mathsfit}{\encodingdefault}{\sfdefault}{m}{sl}
\SetMathAlphabet{\mathsfit}{bold}{\encodingdefault}{\sfdefault}{bx}{n}

\definecolor{cvprblue}{rgb}{0.21,0.49,0.74}
\usepackage[pagebackref,breaklinks,colorlinks,citecolor=cvprblue]{hyperref}
\usepackage{graphicx}
\usepackage{amsmath}
\usepackage{enumitem}
\usepackage[most]{tcolorbox}
\usepackage[utf8]{inputenc} 
\usepackage[T1]{fontenc}    
\usepackage{hyperref}       
\usepackage{booktabs}       
\usepackage{amsfonts}       
\usepackage{nicefrac}       
\usepackage{microtype}      
\usepackage{warpcol}
\usepackage{multirow}
\usepackage{makecell}
\usepackage[misc]{ifsym}
\usepackage{mathrsfs}
\usepackage{xspace}
\usepackage{multicol}
\usepackage{subcaption}
\usepackage{amsthm}
\usepackage[ruled]{algorithm}
\usepackage[noend]{algpseudocode} 
\usepackage{hyperref}
\usepackage[table]{xcolor}
\usepackage{wrapfig}
\usepackage{extarrows}
\usepackage{pifont}
\newtheorem{proposition}{Proposition}
\newtheorem{theorem}{Theorem}

\def\paperID{927} 
\def\confName{CVPR}
\def\confYear{2025}

\title{Theoretical Insights in Model Inversion Robustness and Conditional Entropy Maximization for Collaborative Inference Systems}

\author{
  Song Xia$^{1}$, Yi Yu$^{1}$, Wenhan Yang$^{2}$\thanks{Corresponding Author}, Meiwen Ding$^{1}$, Zhuo Chen$^{2}$,
  \\ Ling-Yu Duan$^{2,3}$, Alex C. Kot$^{1}$, Xudong Jiang$^{1}$ \vspace{1mm} \\
  $^{1}$ROSE Lab, Nanyang Technological University,~$^{2}$Pengcheng Laboratory,~$^{3}$Peking University \\
  {\tt\small \{xias0002,yuyi0010,ding0159,eackot,exdjiang\}@ntu.edu.sg,}{~\tt\small yangwh@pcl.ac.cn,}{~\tt\small lingyu@pku.edu.cn}
}

\begin{document}
\maketitle
\begin{abstract}
By locally encoding raw data into intermediate features, collaborative inference enables end users to leverage powerful deep learning models without exposure of sensitive raw data to cloud servers.
{However, recent studies have revealed that these intermediate features may not sufficiently preserve privacy, as information can be leaked and raw data can be reconstructed via model inversion attacks (MIAs).}
Obfuscation-based methods, such as noise corruption, adversarial representation learning, and information filters, enhance the inversion robustness by obfuscating the task-irrelevant redundancy empirically.
However, methods for quantifying such redundancy remain elusive, and the explicit mathematical relation between this redundancy minimization and inversion robustness enhancement has not yet been established.
To address that, this work first theoretically proves that the conditional entropy of inputs given intermediate features provides a guaranteed lower bound on the reconstruction mean square error (MSE) under any MIA.
Then, we derive a differentiable and solvable measure for bounding this conditional entropy based on the Gaussian mixture estimation and propose a conditional entropy maximization (CEM) algorithm to enhance the inversion robustness. 
Experimental results on four datasets demonstrate the effectiveness and adaptability of our proposed CEM; without compromising feature utility and computing efficiency, plugging the proposed CEM into obfuscation-based defense mechanisms consistently boosts their inversion robustness, achieving average gains ranging from 12.9\% to 48.2\%.
Code is available at \href{https://github.com/xiasong0501/CEM}{https://github.com/xiasong0501/CEM}.
\end{abstract}    
\vspace{-4mm}
\section{Introduction}
\label{sec:intro}
\vspace{-2mm}
Deep neural networks (DNNs), trained on extensive datasets, have demonstrated {outstanding} performance across a growing spectrum of complex applications~\cite{nichol2022glide,kirillov2023segment,li2023blip}.
However, the increasing reliance on deploying these powerful models on cloud platforms, such as ChatGPT-4~\cite{achiam2023gpt}, introduces significant privacy and security concerns, 
{as users may upload data containing sensitive information to cloud servers.}
\begin{figure}[t]
    \centering
\includegraphics[width=0.91\linewidth]{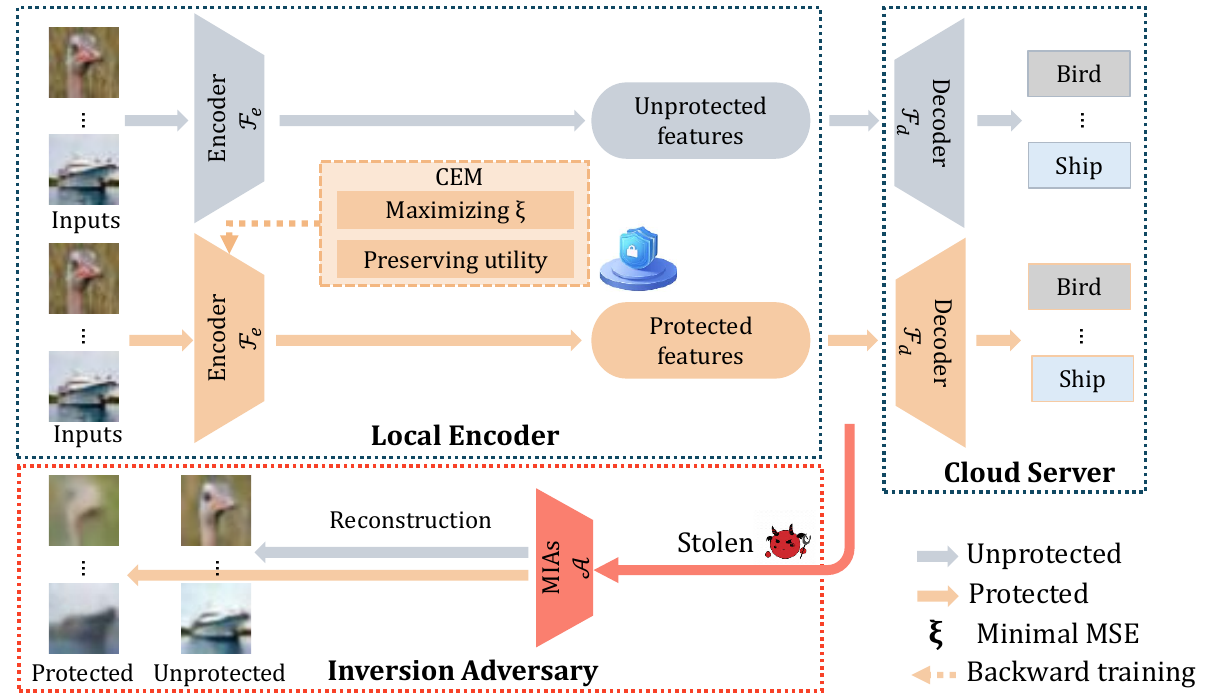} 
    \vspace{-3mm}
    \caption{Privacy protection for collaborative inference via CEM.}
    \label{fig:1 intro}
    \vspace{-7mm}
\end{figure}
Collaborative inference~\cite{shlezinger2021collaborative,thapa2022splitfed,vepakomma2018split} offers a solution by partitioning the deep learning model {across edge devices and cloud servers}, where computations on the initial {shallow} layers are performed locally on the user's device, and only the extracted intermediate features are transmitted to the cloud for subsequent processing. 
This allows end users to utilize powerful neural networks 
{with minimal exposure of their raw inputs, and hence enhances data privacy.}
However, recent works~\cite{he2019model,he2020attacking,carlini2023extracting,mehnaz2022your,struppek2022plug,kariyappa2021maze,sanyal2022towards,nguyen2024label,zong2024ipremover,dibbo2023sok,qiu2024closer,zhao2021exploiting,kong2025pixel,xia2024transferable,yu2024backdoor} have revealed that 
those seemingly minor signals in these intermediate features still contain substantial sensitive information.
As shown in \figureautorefname~\ref{fig:1 intro}, the MIAs can steal those unprotected features to accurately reconstruct the raw inputs.

Existing defense against MIAs can be broadly categorized into cryptography-based~\cite{knott2021crypten,ohrimenko2016oblivious,Pratyush20} and obfuscation-based methods~\cite{jin2024faceobfuscator,gong2023gan,ding2024patrol,ho2024model,yang2023purifier}.
Cryptography-based methods, such as homomorphic encryption~\cite{juvekar2018gazelle,gentry2011implementing} and secure multi-party computation~\cite{wagh2021falcon,Pratyush20}, provide robust theoretical guarantees against MIAs by computing over encrypted data.
However, the inherent computational overhead poses substantial challenges for their scalability on large-scale datasets~\cite{mireshghallah2021not}.
Obfuscation-based defense aims to obfuscate the task-irrelevant redundancy by learning a privacy-preserving feature encoder~\cite{li2022ressfl,li2019deepobfuscator,bertran2019adversarially,xiao2020adversarial,mireshghallah2020shredder,jeong2023noisy,jin2024faceobfuscator,dibbo2024improving}.
%
{Those approaches primarily rely on empirical heuristics, such as assuming a proxy inversion adversary, to estimate task-irrelevant redundancy during encoder optimization. However, a rigorous quantification for evaluating such redundancy remains absent.}
Existing works~\cite{nguyen2023re,yang2022measuring} have indicated that {such} empirical measure is not fully reliable, rendering it insufficient for fully exploring the inversion robustness of the trained feature encoder.
Some methods~\cite{mireshghallah2021not,peng2022bilateral,wang2021improving,maeng2024bounding} employ information-theoretic frameworks to constrain the redundancy.
However, none of them establishes a formal mathematical relationship between information redundancy and robustness against the worst-case inversion adversary, leaving a gap in fully understanding the interplay between redundancy minimization and robustness enhancement.

This work aims to {establish a systematic quantification approach to measure the task-irrelevant yet privacy-critical redundancy} within the intermediate features.
Furthermore, we endeavor to establish a theoretical relationship between the quantified redundancy and the worst-case model inversion robustness, thereby providing a tractable approach to enhance the inversion robustness of existing models against MIAs.
We first demonstrate that the conditional entropy of the input $\vx$ given intermediate feature $\vz$ is strongly correlated with the information leakage, which guarantees a theoretical lower bound on the reconstruction MSE between the original and reconstructed inputs under any inversion adversary.
Moreover, a differentiable and tractable measure is developed for bounding this conditional entropy based on Gaussian mixture estimation.
Utilizing this differentiable measure, we propose a versatile conditional entropy maximization (CEM) algorithm that can be seamlessly plugged into existing empirical obfuscation-based methods, as shown in \figureautorefname~\ref{fig:1 intro}, to consistently enhance their robustness against MIAs. 
The contributions of our work can be summarized as:
\begin{itemize}
    \item We {make the first effort in establishing} a theoretical relationship between the conditional entropy of inputs given intermediate features and the worst-case MIA robustness. Additionally, we derive a differentiable and tractable measure for quantifying this conditional entropy.
    \item Building upon these theoretical insights, we propose a versatile CEM algorithm that can be seamlessly plugged into existing obfuscation defense to enhance its effectiveness in defending against MIAs.
    \item We conduct extensive experiments across four datasets to empirically validate the effectiveness and adaptability of the proposed CEM algorithm. Our findings demonstrate that integrating CEM with existing obfuscation-based defenses consistently yields substantial gains in inversion robustness, without sacrificing feature utility or incurring additional computational overhead.
\end{itemize}


\vspace{-2mm}
\section{Related Work}
\label{sec:related work}
\vspace{-2mm}
\noindent\textbf{Model Inversion Attacks (MIAs).} MIAs represent a serious privacy threat, wherein adversaries reconstruct users' private input data by exploiting the information in model parameters or the redundancy present in intermediate outputs.
The inversion adversaries can leverage a generative model~\cite{zhang2020secret,wang2021variational,yin2023ginver,yuan2023pseudo,nguyen2023re,nguyen2024label}, such as the generative adversarial network (GAN)~\cite{goodfellow2014generative}, or a DNN-based decoder~\cite{yang2019neural,salem2020updates,li2022ressfl} to learn the underlying mapping between intermediate features and original inputs, thereby uncovering the hidden inversion patterns. 
MIAs can be executed under a variety of conditions. 
Early research on MIAs primarily focuses on the white-box setting where the adversary has full access to the models along with the training data~\cite{fredrikson2015model,wang2019beyond,hitaj2017deep,zhang2020secret}.   
However, recent work indicated that only by accessing the intermediate features and with little prior knowledge of the data~\cite{gong2023gan,melis2019exploiting,mehnaz2022your,zong2024ipremover}, the adversary can launch strong MIAs.
%

\noindent\textbf{Obfuscation-based MIAs defense mechanisms.} The obfuscation-based defense methods protect the input privacy by obfuscating the redundant information in the intermediate features.
These methods typically adopt strategies such as perturbing network weights~\cite{abuadbba2020can} or intermediate features~\cite{mireshghallah2020shredder,jeong2023noisy} via noise corruption, purifying intermediate features through frequency domain filtering~\cite{wang2022privacy,mi2023privacy,mi2024privacy} or sparse coding~\cite{jin2024faceobfuscator,dibbo2024improving}, and training inversion-robust encoders via adversarial representation learning~\cite{li2022ressfl,li2019deepobfuscator,bertran2019adversarially,xiao2020adversarial}. 
Those methods generally incur no extra computational overhead during inference, thus serving as a practical and efficient solution for protecting data privacy against MIAs.
Although these approaches offer practical effectiveness in defending against MIAs, they primarily measure such redundancy by some
empirical heuristics without rigorous quantification~\cite{nguyen2023re,yang2022measuring,li2022ressfl}, leading to a sub-optimal trade-off between feature robustness and utility.
Furthermore, a formal mathematical relationship between redundancy and inversion robustness has not been established, leaving a critical gap in comprehensively understanding the interplay between redundancy minimization and robustness enhancement.

\vspace{-2mm}
\section{Methodology}
\vspace{-2mm}
\subsection{Preliminaries}
\vspace{-2mm}
\textbf{Inversion threats on collaborative inference systems:} We consider MIAs on collaborative inference systems, where deep learning models are split into a lightweight encoder $\mathcal{F}_e$ deployed locally and a decoder $\mathcal{F}_d$ deployed in the cloud. 
The end-users first encode their raw input $\vx$ into the intermediate features $\vz=\mathcal{F}_e(\vx)$ locally, and then upload them to the cloud for prediction.
This process is known to be susceptible to MIAs, as the intermediate features $\vz$ {is considered as} a direct representation of the input~\cite{fang2023gifd,titcombe2021practical}.
%
%

\noindent\textbf{The attack model:} 
We consider the scenarios where the inversion adversary $\mathcal{A}$ can steal those user-shared intermediate features $\vz$ to reconstruct raw inputs.
Such threats may arise from the presence of an untrustworthy cloud server or unauthorized access to data traffic during transmission.
%
To evaluate the worst-case inversion robustness, we assume the adversary possesses white-box access to the feature encoder $\mathcal{F}_e$ and has full prior knowledge of the input data space $\mathcal{X}$.
The MIA process is formulated as $\hat{\vx}=\mathcal{A}(\vz,\mathcal{F}_e,\mathcal{X})$.
This information inversion can be achieved by utilizing the DNN-based decoder~\cite{li2022ressfl,han2023reinforcement} or the generative models~\cite{zhang2020secret,yin2023ginver,nguyen2024label} to learn the potential mapping from $\vz \to \vx$.

\subsection{Theoretical Insights on Worst-case Robustness}
\vspace{-2mm}
Assume that the information leakage is quantified by the MSE between the original $\vx$ and the reconstructed $\hat{\vx}$, \textit{i.e.,} $\left\| {\hat \vx - \vx} \right\|_2^2/d$, where $d$ is the input dimensionality.
\vspace{-1mm}
\begin{proposition}[\textbf{Minimal reconstruction MSE $\xi $}]
In the worst-case scenarios, where the adversary $\mathcal{A}(\vz,\mathcal{X},\mathcal{F}_e)$ precisely estimates the posterior probability $\mathbb{P}(\vx | \vz)$ based on the extensive data prior $\mathcal{X}$ and the white-box access to the feature encoder $\mathcal{F}_e$, the expectation of the minimal reconstruction MSE $\xi $ over the whole dataset satisfies that:
\vspace{-2mm}
\begin{equation} \small
{\xi }\cdot d  = \mathbb{E}_\mathcal{Z}\mathbb{E}_\mathcal{X}\left[\left\| {\vx- \mathbb{E}\left[\vx|\vz\right]} \right\|_2^2 | \vz\right]=\mathbb{E}_\mathcal{Z}\left[Tr(Cov(\vx| \vz)\right].
    \label{eq:opt_inverattack}
\end{equation}
    \label{proposition1}
\end{proposition}
\vspace{-7mm}

\noindent $Cov(\vx|\vz)$ is the covariance matrix of $\vx$ conditioned on $\vz$, and $Tr$ is the trace operator.
Eq.~\ref{eq:opt_inverattack} is established based on the fact that the minimized MSE is given by the expectation of the posterior probability of $\vx$ given $\vz$.
\vspace{-1mm}
\begin{theorem}[\textbf{Lower bound on the minimal reconstruction MSE $\xi$}]
Let $\mathcal{H}(\vx| \vz)$ denote the conditional entropy of the input $\vx$ given the intermediate feature $\vz$. The minimal reconstruction MSE $\xi$ is bounded by: $\xi \ge \frac{1}{(2\pi e)}exp(\frac{2\mathcal{H}({\vx}|{\vz})}{d})$.
\label{theorem1}
\end{theorem}
\vspace{-5mm}

\noindent Theorem~\ref{theorem1} provides a lower bound on the minimal reconstruction MSE $\xi$ in terms of the conditional entropy $\mathcal{H}({\vx}|{\vz})$, which serves as a robust measure for the worst-case robustness against MIAs.
The reconstruction MSE $\xi$ and the conditional entropy are highly correlated.
Physically, a higher $\mathcal{H}(\vx | \vz)$ means a more uncertain estimation of $\vx$ given $\vz$, resulting in a larger estimated error $\xi$.
Thus, indicated by Theorem~\ref{theorem1}, a straightforward way to enhance the inversion robustness is to maximize the $\mathcal{H}({\vx}|{\vz})$ during training.
The proof of Theorem~\ref{theorem1} is in the supplementary materials S.1.

However, deriving a closed-form expression between $\vz$ and $\vx$ is exceedingly difficult due to the inherent intractability of neural networks, especially when applied to large-scale datasets.
This renders the direct calculation of $\mathcal{H}({\vx}|{\vz})$ computationally prohibitive, making its maximization via backpropagation to eliminate the information redundancy during training exceedingly intractable.
To solve this, a differentiable and computationally efficient lower bound on the conditional entropy $\mathcal{H}(\vx|\vz)$ is introduced in the next section to facilitate its maximization during training.

\subsection{Differentiable Bound on Conditional Entropy}
\vspace{-2mm}
\begin{proposition}[\textbf{The conditional entropy under uncertain encoding}]
Without loss of the generality, we consider that the encoding process $\vx \to \vz$ consists of two components: a deterministic mapping: $\vx \to \hat{\vz}$ and a stochastic process: $\vz=\hat{\vz}+\varepsilon$, where $\varepsilon$ is a random noise. $\mathcal{H}(\vx|\vz)$ satisfies:
\vspace{-1mm}
\begin{equation}\small    
\mathcal{H}(\vx|\vz)={\mathcal{H}(\vx)-\mathcal{H}(\vz)}+{\mathcal{H}(\vz|\hat{\vz})}= \mathcal{H}(\vx)-\mathcal{I}(\vz;\hat{\vz}).
\label{eq:enc_condition_entropy}
\end{equation}
\label{proposition3}
\end{proposition}
\vspace{-8mm}

\noindent Proposition~\ref{proposition3} is derived based on the definition of joint entropy, where $\mathcal{H}(\vx|\vz)= \mathcal{H}(\vx)+\mathcal{H}(\vz|\vx)-\mathcal{H}(\vz)$.
The equation $\mathcal{H}(\vz|\vx)=\mathcal{H}(\vz|\hat{\vz})$ holds due to the causal dependency from $\vx \to \hat{\vz} \to \vz$ and $\vx \to \hat{\vz}$ is deterministic.
Since $\mathcal{H}(\vx)$ remains constant for a given data distribution, the primary challenge in maximizing the conditional entropy $\mathcal{H}(\vx|\vz)$ indicated by Eq.~\ref{eq:enc_condition_entropy} reduces to deriving a tractable and differentiable measure for $\mathcal{I}(\vz;{\hat{\vz}})$.

\begin{figure*}[tbh]
    \centering  \includegraphics[width=0.90\textwidth]{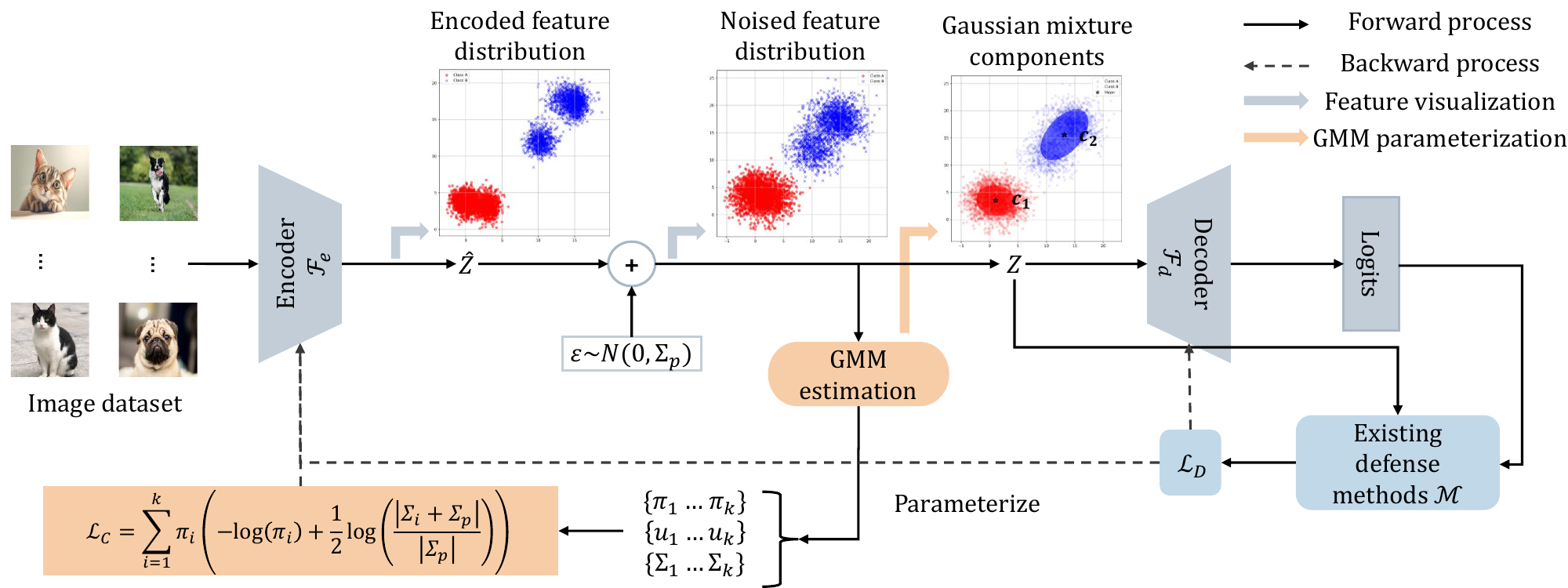} 
    \vspace{-4mm}
    \caption{The versatile conditional entropy maximization algorithm for collaborative learning systems with split encoder and decoder.}
    \label{fig:1 flowchat}
    \vspace{-4mm}
\end{figure*}

Consider the scenario where the task has $n$ discrete targets $\vy =\{y_1,..., y_n \}$.
Following previous statements, we consider that the encoding process can be separated into a deterministic mapping: $\vx \to \hat \vz$ and a noise corruption process: $\vz= \hat \vz + \varepsilon$, where $\varepsilon \sim \mathcal{N}(0,\Sigma_p)$ is a Gaussian noise. 
We assume that the distribution of the encoded feature $\hat{\vz}$ can be effectively estimated by a $k$-component Gaussian mixture distribution denoted as ${\hat \vz} \sim\sum_{i=1}^{k}\pi_i\mathcal{N}(\mu_i,\Sigma_i)$, leveraging the proven capability of Gaussian mixtures to closely approximate complex natural data distributions~\cite{reynolds2009gaussian}.
\vspace{-1mm}
\begin{theorem}[\textbf{Differentiable lower bound on $\mathcal{H}(\vx|\vz)$}]
Given $\hat{\vz}$ follows the Gaussian mixture distribution and $\varepsilon$ is a Gaussian noise. The encoded feature $\vz=\hat{\vz}+\varepsilon$ also follows a Gaussian mixture distribution with $\vz \sim\sum_{i=1}^{k}\pi_i\mathcal{N}(\mu_i,\Sigma_i+\Sigma_p)$. Consequently, the conditional entropy $\mathcal{H}(\vx|{{\vz}})$ is bounded by:
\vspace{-3mm}
\begin{equation}\small
\mathcal{H}(\vx|{{\vz}})\!\ge\! \mathcal{H}(\vx) \!-\! \sum_{i=1}^{k} \!\pi_i\!\left( \!-\!\log(\pi_i) 
\!+\! \frac{1}{2}\log\left(\frac{ \left| \Sigma_i\!+\!\Sigma_p\right|}{\left|\Sigma_p\right|} \right) \right). 
\label{eq:MI_upperbound}
\end{equation}
\vspace{-5mm}
\label{theorem2}
\end{theorem}
%
\noindent Theorem~\ref{theorem2} offers an effective and efficient way to bound the conditional entropy $\mathcal{H}(\vx|{{\vz}})$, as the parameters $\pi_i$ and $\Sigma_i$ are all tractable and differentiable with respect to $\hat{\vz}$.
Given the extracted $\hat{\vz}$ after deterministic mapping and the Gaussian noise $\varepsilon$, one can easily derive the parameters $\pi_i$ and $\Sigma_i$ using a Gaussian mixture model.
This facilitates the maximization of $\mathcal{H}(\vx|{{\vz}})$ via gradient backpropagation utilizing Eq.~\ref{eq:MI_upperbound}, thereby improving the worst-case inversion robustness of the trained model.
The proof of the Theorem~\ref{theorem2} can be found in the supplementary material S.2.  

\subsection{Practical Insights on the Utility and Robustness Trade-off}
\label{trade-off}
\vspace{-1mm}
\textbf{Feature utility}: Building on the previous discussion, the encoded feature $\vz$ follows a Gaussian mixture distribution with $\vz \sim\sum_{i=1}^{k}\pi_i\mathcal{N}(\mu_i,\Sigma_i+\Sigma_p)$.
In general, we assume that each Gaussian component or cluster belongs to one specific target. 
Ideally, we have $k=n$ to accomplish the task.
Due to the causal dependency from $\vx \to \hat{\vz} \to \vz$, the utility of the intermediate feature $\vz$ is intrinsically linked to its correlation with preceding states.
Thereby, the utility of $\vz$ is positively correlated with the expected posterior probability that $\hat \vz$ belonging to its original cluster $j$ given $\vz$, which is:
\vspace{-1mm}
\begin{equation}\small
\mathbb{E}_\mathcal{Z}\left[ \mathbb{P}(({\hat \vz} \in j|\vz))\right]=\mathbb{E}_\mathcal{Z}\left[ \frac{\pi_j\mathcal{N}(\vz;\mu_j,\Sigma_j+\Sigma_p)}{\sum_{i=1}^{k}\pi_i\mathcal{N}(\vz;\mu_i,\Sigma_i+\Sigma_p)} \right].
    \label{eq:opt_pre_acc}
\end{equation}

Combing Eq.~\ref{eq:MI_upperbound} and Eq.~\ref{eq:opt_pre_acc}, we can get that the utility and robustness of the feature $\vz$ are dependent on the $\mathbb{E}_\mathcal{Z}\left[ \mathbb{P}(({\hat \vz}=j|\vz))\right]$ and the $\mathcal{I}(\vz;{\hat{\vz}})$ respectively.
These two items are empirically closely linked, leading to a trade-off between utility and robustness.
Additionally, the following practical insights can be derived:
\begin{itemize}
    \item \textbf{Physical interpretation of $\mu_i$ and $\Sigma_i$}: Physically, the mean $\mu_i$ generally captures dominant or representative characteristics of a cluster, while the covariance reflects the degree of variability or noise within the cluster. Consistent with this interpretation, our analysis reveals that $\mu_i$ is crucial for evaluating the feature utility. In contrast, the covariance $\Sigma_i$ is closely linked to the task-irrelevant redundancy in $\hat{\vz}$, which diminishes feature utility and increases susceptibility to MIAs.
    \item \textbf{Impact of noise corruption on inversion robustness}: The noise corruption introduced by $\Sigma_p$ perturbs both the task-important and redundant information, thus balancing the trade-off between utility and robustness. When there is a distinctive difference between the dominant features, such as $\min_{i\ne j} \left \| \mu_i -\mu_j  \right \|_2^2 \gg Tr(\Sigma_p) \gg Tr(\Sigma_i+\Sigma_j)$, adding the noise greatly corrupts the redundancy while bringing a small influence on the utility.
    \item \textbf{Ideal situation}: For the ideal situation, we have $\Sigma_i=\Sigma_p = \textbf{0}$, $k=n$ and, ${\pi}_i=\mathcal{P}(y_i)$. Thereby, we can easily get $\mathbb{E}_\mathcal{Z}\left[ \mathbb{P}(({\hat \vz}=j|\vz))\right]=1$ and ${\mathcal{H}(\vz)=\mathcal{H}(\vy)}$, indicating that the feature $\vz$ contains only the necessary information for the task, without any redundancy.
\end{itemize}

\noindent However, for a lightweight encoder deployed on edge devices, it is exceedingly challenging to diminish all redundancies while preserving only the essential features.
An alternative approach for developing a robust feature encoder is to jointly optimize both the covariance matrices $\Sigma_i$ and the distance between the $\mu_j$ among different clusters, leading to $\min_{i\ne j} \left \| \mu_i -\mu_j  \right \|_2^2 \gg Tr(\Sigma_p) \gg Tr(\Sigma_i+\Sigma_j)$.
Thus, incorporating appropriate noise can effectively diminish the residual redundancies, achieving the maximization of the lower bound on $\mathcal{H}(\vx|\vz)$ while ensuring prediction accuracy.

\begin{figure}[t]
\vspace{-4mm}
\begin{algorithm}[H]\small
    \caption{Conditional entropy maximization algorithm}
    \begin{algorithmic}[1]
    \State \textbf{Input:} local encoder $\mathcal{F}_e$, cloud decoder $\mathcal{F}_d$, image dataset $\mathcal{D}$ of size $N$, training epochs $m$, conditional entropy loss $\mathcal{L}_C$ with weight factor $\lambda$, Gaussian noise $\varepsilon$, existing defense methods $\mathcal{M}$ with loss $\mathcal{L}_D$, Gaussian mixture model $GMM$, number of mixture clusters $k$.
    \State \text{\textbf{Initialize} $\mathcal{F}_e$, $\mathcal{F}_d$}
    \For{epoch from $1$ to $m$}
    \State $\vz \gets \text{concatenate}\{\mathcal{F}_e(\vx)+\varepsilon, \forall \vx \in \mathcal{D}\}$
    \State Estimate the distribution of $\vz$ as a $k$ component Gaussian mixture :$\sum_{j=1}^{k}\pi_j\mathcal{N}(\mu_j,\Sigma_j+\Sigma_p)$ using $GMM(\vz, k)$
    \For{image batch $(\vx_i, \vy_i)$ in the dataset $\mathcal{D}$}
    \State ${\vz}_i = \mathcal{F}_e(\vx_i)+\varepsilon_i$
    \State ${\vy}'_i=\mathcal{F}_d(\vz_i)$
    \State Assign ${\vz}_i$ to the nearest cluster
    \For{cluster $j$ $\leftarrow$ 1 to $k$}
    \State Update the weight $\pi_j$ by Eq.~\ref{eq:upd_weights}
    \State Update the covariance matrix by Eq.~\ref{eq:upd_convariance}
    \EndFor
    \State {Calculate} $\mathcal{L}_D$ by $\mathcal{M}(\vx_i,\vy_i,\vz_i,{\vy}'_i$)
    \State {Calculate} $\mathcal{L}_C$ by 
    Eq.~\ref{eq:loss_conditional entropy}.
    \State Optimize $\min_{\mathcal{F}_e,\mathcal{F}_d} \left( \mathcal{L}_D + \lambda*\mathcal{L}_C \right)$
    \EndFor
    \EndFor
    \State \textbf{Output:} Trained encoder $\mathcal{F}_e$ and decoder $\mathcal{F}_d$
    \end{algorithmic}
    \label{alg1} 
\end{algorithm}
\vspace{-8mm}
\end{figure}

\section{The Versatile Conditional Entropy Maximization Algorithm for Collaborative Inference}
\vspace{-1mm}
Building upon the derived theoretical insights on the differentiable lower bound of the conditional entropy $\mathcal{H}(\vx|\vz)$ and the practical insights on the utility and robustness trade-off, this section introduces a versatile conditional entropy maximization (CEM) algorithm. 
The CEM algorithm is designed to enhance the robustness of existing obfuscation-based defense mechanisms against MIAs in collaborative inference systems.
For a given defense mechanism, denoted as $\mathcal{M}$, the proposed CEM algorithm begins by assessing its worst-case robustness. 
This is achieved through a Gaussian mixture estimation process on the feature ${\vz}$ to determine the parameters $\mu_i$ and $\Sigma_i$. 
Subsequently, during each training batch, the algorithm maximizes the lower bound of $\mathcal{H}(\vx|\vz)$ specified in Theorem~\ref{theorem2} to strengthen robustness against MIAs.
The details of the proposed CEM algorithm are presented in \figureautorefname~\ref{fig:1 flowchat} and Algorithm~\ref{alg1}.

\textbf{Key techniques:} Let $\mathcal{F}_e$ denote the local feature encoder and $\mathcal{F}_d$ denote the decoder deployed in the cloud. 
By estimating the distribution of noisy representation $\vz$ with a $k$-component Gaussian mixture $\sum_{i=1}^{k} \pi_i \mathcal{N}(\mu_i, \Sigma_i + \Sigma_p)$, the lower bound of the conditional entropy $\mathcal{H}(\vx|\vz)$ can be derived based on  Eq.~\ref{eq:MI_upperbound}, utilizing the covariance matrices $\Sigma_i$ and the weights of the mixture components $\pi_i$.
Since both of them are differentiable and solvable to the feature representation $\vz$, this lower bound of the conditional entropy $\mathcal{H}(\vx|\vz)$ can be effectively maximized via the gradient back-propagation.
As described in Algorithm~\ref{alg1}, each training iteration begins by fitting the distribution of $\vz$ with a Gaussian mixture model (GMM). 
This involves optimizing the component weights $\pi$, means $\mu$, and covariance matrices $\Sigma$ to minimize the estimation error. 
For each data batch, the extracted feature representations are assigned to the mixture component $j$ corresponding to the nearest mean $\mu_j$, with the mixture parameters updated accordingly.
The component weights $\pi_j$ are updated by:
\vspace{-1mm}
\begin{equation}\small
\pi_j=\frac{\pi_j(N-N_{batch})+n_j}{N},
\label{eq:upd_weights}
\end{equation}
\noindent where $N$ is the total length of the dataset and $N_{batch}$ is the batch size. $n_j$ is the number of feature representations that are assigned to the mixture component $j$. The covariance matrices are updated by:
\begin{equation}\small
\Sigma_j = \left(1 - \frac{n_j}{\pi_j N}\right)\Sigma_j + \frac{n_j}{\pi_j N} \Delta \Sigma_j,
\label{eq:upd_convariance}
\vspace{-1mm}
\end{equation}
\noindent where $\Delta \Sigma_j$ is the calculated covariance based on newly assigned feature representations. 
The updated parameters are then utilized to compute the conditional entropy loss $\mathcal{L}_C$, defined as:
\vspace{-2mm}
\begin{equation}\small
\mathcal{L}_C=\sum_{i=j}^{k} \pi_j\left( -\log(\pi_j) 
+ \frac{1}{2}\log\left(\frac{ \left| \Sigma_j+\Sigma_p\right|}{\left|\Sigma_p\right|} \right) \right).
\label{eq:loss_conditional entropy}
\end{equation}
\noindent 
This loss function is then utilized as an effective indicator of worst-case inversion robustness and is minimized through gradient-based optimization.

\textbf{The combination with other defense strategies:}
As shown in \figureautorefname~\ref{fig:1 flowchat}, existing obfuscation-based defense methods, such as noise corruption~\cite{jeong2023noisy,titcombe2021practical}, adversarial representation learning~\cite{li2022ressfl,ding2024patrol}, and feature purification~\cite{jin2024faceobfuscator,dibbo2024improving,vepakomma2020nopeek} are collectively denoted as $\mathcal{M}$, and an auxiliary training loss $\mathcal{L}_D$ is used to encapsulate their impact in the backward optimization process.
For approaches that involve architectural modifications, such as introducing information filtering layers or adversarial reconstruction models, we adhere to their defined configurations to incorporate our method.
The proposed CEM algorithm can be seamlessly plugged into other defense mechanisms by embedding Gaussian mixture estimation within the training process and optimizing the combined loss function:
\vspace{-1mm}
\begin{equation}\small
\mathcal{L}=\mathcal{L}_D+\lambda*\mathcal{L}_C,
\label{eq:joint_loss}
\vspace{-1mm}
\end{equation}
\noindent where $\lambda$ is a weight factor to balance the optimization. The utilization of the proposed CEM algorithm allows for an effective evaluation of worst-case inversion robustness of the feature representation throughout training and further optimizes this robustness via gradient backward propagation.

\begin{figure}[t]
    \centering
\includegraphics[width=0.80\linewidth]{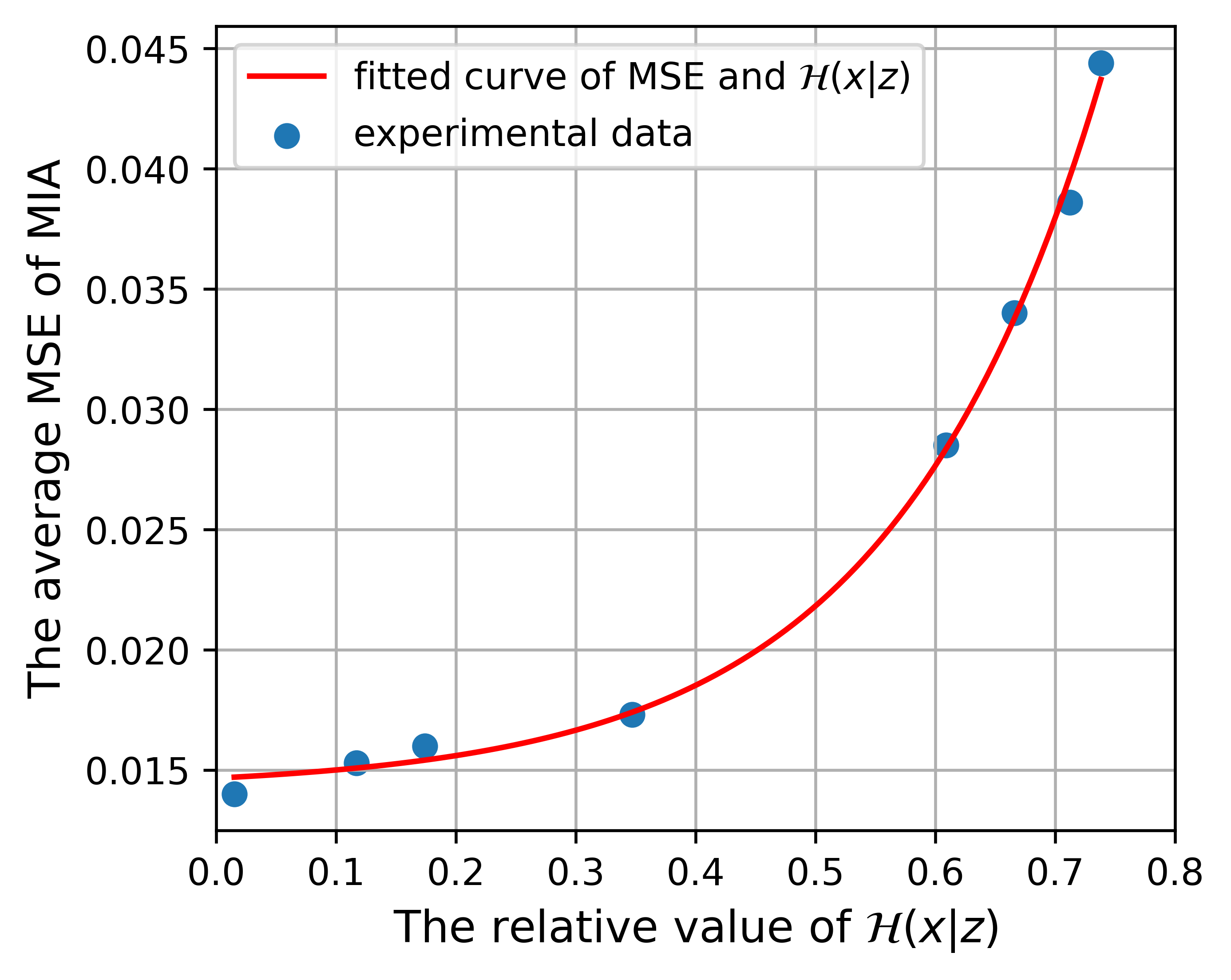} 
    \vspace{-5mm}
    \caption{Reconstruction MSE Vs. $\mathcal{H}(\vx|\vz)$ on the CIFAR10.}
    \label{fig:2 relation of MSE and entropy}
    \vspace{-6mm}
\end{figure}
\textbf{Conditional entropy $\mathcal{H}(\vx|\vz)$ and reconstruction MSE by MIAs in practical scenarios:} Theorem~\ref{theorem1} indicates an exponential relationship between the lower bound on minimal MSE $\xi$ and the conditional entropy $\mathcal{H}(\vx|\vz)$.
We thereby empirically evaluate this relationship on the CIFAR10 dataset by training multiple models with varying levels of measured $\mathcal{H}(\vx|\vz)$ and assessing their inversion robustness by assuming a proxy inversion adversary. 
Following the experimental setting in~\cite{li2022ressfl}, we utilize a VGG11 model with the first 2 convolutional layers as encoder $\mathcal{F}_e$ and the rest of layers as the decoder $\mathcal{F}_d$.
We fix the $\lambda=16$ and train different models using different Gaussian noise with variance varying from 0.01 to 0.3, thus leading to varying $\mathcal{H}(\vx|\vz)$. 
The relative conditional entropy $\mathcal{H}(\vx|\vz)$ is calculated by Eq.~\ref{eq:MI_upperbound} (we use 'relative' due to $\mathcal{H}(\vx)$ is constant and its calculation over the whole data space is intractable. So a proper value is selected to replace it).  
We report the average reconstruction MSE by training a DNN-based MIA decoder $\mathcal{A}(\vz,\mathcal{X},\mathcal{F}_e)$ with the same architecture as~\cite{li2022ressfl} and allow its full access to the training data and encoder $\mathcal{F}_e$.
The analytic results are shown in \figureautorefname~\ref{fig:2 relation of MSE and entropy}, revealing a strong exponential correlation between the $\mathcal{H}(\vx|\vz)$ and the MIA reconstruction MSE.

\section{Experimental Results}
\label{sec:Experimental results}
\vspace{-1mm}
\subsection{Experimental Settings}
\vspace{-1mm}
\setlength{\tabcolsep}{8pt}
\begin{table}[t]
    \centering
    \caption{Comparative results on CIFAR10 dataset: we present the prediction accuracy and reconstruction MSE of different defense methods using the VGG11 model before and after the integration with the proposed CEM algorithm. The last 2 rows report the average performance of all methods, comparing results with and without incorporating the CEM algorithm.}
    \vspace{-3mm}
    \resizebox{0.47\textwidth}{!}{%
    \begin{tabular}{lccccc}
        \toprule
        \multirow{2}{*}{\textbf{Methods}} &\multirow{2}{*}{\textbf{Acc.$\uparrow$}}& \multicolumn{2}{c}{\textbf{Dec.-based MSE~\cite{li2022ressfl}}} & \multicolumn{2}{c}{\textbf{GAN-based MSE~\cite{zhang2020secret}}} 
        \\
        \cmidrule(lr){3-4} \cmidrule(lr){5-6} 
        &  & \textbf{Train$\uparrow$} &\textbf{Infer$\uparrow$} & \textbf{Train$\uparrow$} &\textbf{Infer$\uparrow$}
        \\
        \midrule
        No\_defense&91.86 &0.0013&0.0014&0.0014&0.0016
        \\
        \midrule
        {Bottleneck} 
        & 90.87 & 0.0036 & 0.0041 & 0.0039 & 0.0045\\
        Bottleneck+CEM& 90.69 & 0.0054 & 0.0058 & 0.0076 &0.0080
        \\ \midrule
        DistCorr~\cite{vepakomma2020nopeek} &89.52& 0.0074 & 0.0082 & 0.0079 &0.0088 
        \\
        DistCorr+CEM & 89.80 & 0.0090 & 0.0093 &0.0094&0.0096 \\
        \midrule
        Dropout~\cite{he2020attacking} 
        & 87.75 & 0.0098 & 0.0104 & 0.0099 & 0.0111\\
        Dropout+CEM& 87.53 & 0.0129 & 0.0134 & 0.0132 & 0.0142\\ 
        \midrule
        PATROL~\cite{ding2024patrol} &89.58& 0.0245& 0.0293 & 0.0257 &0.0307 \\
        PATROL+CEM& 89.67 & 0.0304 & 0.0335 &0.0313 &0.0347 \\
        \midrule
        ResSFL~\cite{li2022ressfl} &89.68& 0.0146 & 0.0240 & 0.0158 &0.0243\\
        ResSFL+CEM&90.11 & 0.0201& 0.0251 &0.0229&0.0273  \\
        \midrule
        Noise\_Nopeek~\cite{titcombe2021practical}
        & 87.19 & 0.0152 & 0.0159 & 0.0156 & 0.0165 \\
        Noise\_Nopeek+CEM& 87.08 & 0.0174 & 0.0174 & 0.0173 &0.0178\\
        \midrule
        Noise\_ARL~\cite{jeong2023noisy} &87.78& 0.0290 &0.0336 & 0.0304 &0.0342
        \\
        Noise\_ARL+CEM& 87.62 & 0.0330 & 0.0355 &0.0341 &0.0363\\
        \midrule
        \rowcolor{gray!20}
        \textbf{Average w/o CEM}
        & 88.91 & 0.0148 & 0.0179 & 0.0156 & 0.0185\\
        \rowcolor{gray!20} \textbf{Average w/ CEM} &\textbf{88.92} & \textbf{0.0183} & \textbf{0.0200} & \textbf{0.0194}  & \textbf{0.0211}\\
        \bottomrule
    \end{tabular}}
    \vspace{-5mm}
    \label{tab:1}
\end{table}
We investigate the inversion robustness of collaborative inference models implemented on three general object classification datasets: CIFAR-10~\cite{krizhevsky2009learning}, CIFAR-100~\cite{krizhevsky2009learning}, TinyImageNet~\cite{le2015tiny}, and a face recognition dataset FaceScrub~\cite{ng2014data}.
For the CIFAR-10, CIFAR-100, and FaceScrub datasets, we use the VGG11 as the basic model and for the TinyImageNet dataset, we use the ResNet-20 as the basic model.

\vspace{0.3mm}
\noindent\textbf{Collaborative inference system:} 
To simulate the collaborative inference, we split the VGG11 model by allocating the first two convolutional layers as the local encoder and assigning the remaining layers as the cloud decoder.
For the ResNet-20 model, the first four convolutional years are allocated as the local encoder, with the remaining blocks functioning as the cloud-based decoder. 
This partitioning yields a computational distribution where the encoder is responsible for approximately 10\% of the total computation and the decoder handles the remaining 90\%.
Unless otherwise specified, we maintain the same partitioning scheme across all evaluated methods for a fair comparison.
We utilize the Stochastic Gradient Descent (SGD) optimizer to jointly optimize both the encoder and decoder with an initial learning rate of 0.05 and an appropriate learning rate decay.
The VGG11 model is trained over 240 epochs and the ResNet20 model is trained over 120 epochs. 

\vspace{0.3mm}
\noindent\textbf{Obfuscation defense methods:} The proposed CEM can be flexibly integrated into the model training process to enhance the worst-case inversion robustness by introducing an auxiliary Gaussian mixture estimation to maximize the conditional entropy.
We evaluate the effectiveness of the proposed CEM in enhancing several existing defense methods, including the information pruning-based methods called DistCorr~\cite{vepakomma2020nopeek} and Dropout~\cite{he2020attacking}; adversarial representation learning-based method called ResSFL~\cite{li2022ressfl} and PATROL~\cite{ding2024patrol}; and noise corruption-based method called Noise\_Nopeek~\cite{titcombe2021practical} and Noise\_ARL~\cite{jeong2023noisy}.
As observed in~\cite{li2022ressfl}, the integration of bottleneck layers substantially enhances inversion robustness.
To ensure a fair and rigorous comparison, we incorporate a bottleneck layer with 8 channels for both pre- and post-processing of extracted feature representations across all evaluated methods.
For PATROL~\cite{ding2024patrol} that introduces additional layers within the encoder and maintains the inference efficiency through network pruning techniques, we adjust the pruning ratio to ensure the encoder in PATROL remains the same size as other methods.
To analyze the effect of the proposed CEM algorithm, we evaluate the performance of all methods both before and after integrating them with the CEM algorithm.
If not otherwise stated, we set hyperparameters $\lambda=16$ and utilize the isotropic Gaussian noise $\varepsilon$ with a standard deviation of 0.025. 
We set the number of Gaussian mixture components $k=3n$ due to the limited number of training data and the imperfect representation ability of the shallow encoder.

\noindent\textbf{Threat model:} To evaluate the worst-case inversion robustness, we consider the white-box scenarios where the MIAs are trained with full access to the collaborative inference model and training dataset. 
We split the training and testing data strictly following the default setting, \eg 50,000 images for training and 10,000 for testing on the CIFAR10 dataset.
Two types of inversion attacks: the DNN-based decoding method~\cite{li2022ressfl} and the generative-based inversion method~\cite{zhang2020secret} are utilized as the MIAs.
The detailed architecture and training mechanism of those threat models can be found in the supplementary material S.3.

\noindent\textbf{Evaluation metrics:}
We report the prediction accuracy and the reconstruction MSE to evaluate the utility and inversion robustness of the intermediate feature.
We also report other metrics such as PSNR and SSIM to evaluate the robustness and present the experimental results in the supplementary material S.4.
We consider the information leakage on training and inference data, where we report the MSE on reconstructing the training and testing dataset by MIAs. 
To further explore the intrinsic trade-off between the feature utility and inversion robustness for different defense mechanisms, we present the accuracy-MSE curve by varying the defense hyperparameters, such as the noise strength and CEM loss weight factor $\lambda$. 
\begin{table}[t]
    \centering
    \caption{Comparative results on TinyImageNet dataset: we present the prediction accuracy and reconstruction MSE of different defense methods using the ResNet-20 model with and without integrating the proposed CEM.}
    \vspace{-2mm}
    \resizebox{0.48\textwidth}{!}{%
    \begin{tabular}{lccccc}
        \toprule
        \multirow{2}{*}{\textbf{Methods}} &\multirow{2}{*}{\textbf{Acc.$\uparrow$}}& \multicolumn{2}{c}{\textbf{Dec.-based MSE~\cite{li2022ressfl}}} & \multicolumn{2}{c}{\textbf{GAN-based MSE~\cite{zhang2020secret}}}
        \\
        \cmidrule(lr){3-4} \cmidrule(lr){5-6} 
        &  & \textbf{Train$\uparrow$} &\textbf{Infer$\uparrow$} & \textbf{Train$\uparrow$} &\textbf{Infer$\uparrow$}
        \\
        \midrule
        No\_defense&53.73&0.0025&0.0022&0.0020&0.0021
        \\
        \midrule
        {Bottleneck} 
        & 52.77 & 0.0107 & 0.0103 & 0.0091 & 0.0092\\
        Bottleneck+CEM& 52.25& 0.0140 &0.0136 & 0.0123 &0.0125\\
        \midrule
        DistCorr~\cite{vepakomma2020nopeek} &51.79& 0.0148 & 0.0145 & 0.0126	 &0.0136	 \\
        DistCorr+CEM & 51.78 & 0.0195 & 0.0190 &0.0167 &0.0168 \\
        \midrule
        Dropout~\cite{he2020attacking} 
        & 50.72 & 0.0165 & 0.0163 & 0.0148 & 0.0151\\
        Dropout+CEM& 50.75 & 0.0188 & 0.0186 & 0.0167 & 0.0172\\
        \midrule
        PATROL~\cite{ding2024patrol} &51.75& 0.0187& 0.0187 & 0.0168 &0.0176\\
        PATROL+CEM& 51.64 & 0.0211 & 0.0209 & 0.0185&0.0189	 \\
        \midrule
        ResSFL~\cite{li2022ressfl} &51.99& 0.0173 & 0.0172 & 0.0157 &0.0161\\
        ResSFL+CEM&52.07 & 0.0197& 0.0194 &0.0180&0.0173  \\
        \midrule
  
        Noise\_Nopeek~\cite{titcombe2021practical}
        & 51.63 & 0.0161 & 0.0157 & 0.0147 & 0.0152\\
        Noise\_Nopeek+CEM& 52.01 & 0.0183 & 0.0180 & 0.0166 &0.0167 \\
        \midrule
        Noise\_ARL~\cite{jeong2023noisy} &51.03 & 0.0229 &0.0224 & 0.0204 &0.0205 \\
        Noise\_ARL+CEM& 50.85 & 0.0281 & 0.0271 &0.0251 &0.0253\\
        \midrule
        \rowcolor{gray!20}
        \textbf{Average w/o CEM}
        & \textbf{51.66} & 0.0167 & 0.0164 & 0.0148 & 0.0153\\
        \rowcolor{gray!20} \textbf{Average w/ CEM} &51.62 & \textbf{0.0199} & \textbf{0.0195} & \textbf{0.0177} &\textbf{0.0178}
        \\
        \bottomrule
    \end{tabular}}
    \vspace{-5mm}
    \label{tab:2}
\end{table}
\subsection{Results on Different Datasets}
\vspace{-1mm}
\noindent \textbf{Results on CIFAR10:}
The performance of different defense methods on the CIFAR10 dataset is presented in \tableautorefname~\ref{tab:1}.
We provide a comparative analysis of each method, both before and after integrating the proposed CEM algorithm.
The results indicate that the features, extracted by the lightweight encoder without defense mechanisms, exhibit significant redundancy, which can be exploited by MIAs to reconstruct high-fidelity inputs. 
Information pruning methods, such as DistCorr~\cite{vepakomma2020nopeek} and dropout~\cite{he2020attacking} that are based on some empirical observations, inevitably remove task-specific information when eliminating the redundancy, leading to a degradation of the prediction accuracy. 
Methods based on adversarial representation learning (ARL), such as ResSFL, PATROL, and Noise\_ARL, offer a better trade-off in feature utility and robustness, achieving substantial robustness gain with a moderate accuracy drop.
By incorporating a Gaussian mixture estimation process that is independent of the prediction process, the proposed CEM algorithm effectively increases the lower bound of the reconstruction MSE via maximizing the conditional entropy.
This positions the CEM algorithm as a highly adaptable framework for assessing and enhancing the inversion robustness of a wide range of collaborative inference models.
The results in \tableautorefname~\ref{tab:1} show that the integration of the CEM algorithm consistently enhances all defense methods, \textbf{yielding an average increase in reconstruction MSE of 24.0\% for training data and 12.9\% for inference data,} without compromising prediction accuracy.

\begin{table}[t]
    \centering
    \caption{Comparative results on Facescrub dataset: we present the prediction accuracy and reconstruction MSE of different defense methods using the VGG11 model with and without integrating the proposed CEM. }
    \vspace{-2mm}
    \resizebox{0.48\textwidth}{!}{
    \begin{tabular}{lcccccc}
        \toprule
        \multirow{2}{*}{\textbf{Methods}} &\multirow{2}{*}{\textbf{Acc.$\uparrow$}}& \multicolumn{2}{c}{\textbf{Dec.-based MSE~\cite{li2022ressfl}}} & \multicolumn{2}{c}{\textbf{GAN-based MSE~\cite{zhang2020secret}}}
        \\
        \cmidrule(lr){3-4} \cmidrule(lr){5-6} 
        &  & \textbf{Train$\uparrow$} &\textbf{Infer$\uparrow$} & \textbf{Train$\uparrow$} &\textbf{Infer$\uparrow$}\\
        \midrule
        No\_defense&86.69&0.0012&0.0011&0.0012&0.0014
        \\
        \midrule
        {Bottleneck} 
        & 85.12 & 0.0025 & 0.0025 & 0.0026 & 0.0027\\
        Bottleneck+CEM& 85.00 & 0.0036 & 0.0035 & 0.0038 &0.0037
        \\
        \midrule
        DistCorr~\cite{vepakomma2020nopeek} &83.42& 0.0038 & 0.0038 & 0.0041 & 0.0041 \\
        DistCorr+CEM & 83.78 & 0.0048 & 0.0047 &0.0069 &0.0074 \\
        \midrule
        Dropout~\cite{he2020attacking} 
        & 79.30 & 0.0052 & 0.0052 & 0.0054 & 0.0057\\
        Dropout+CEM& 79.19 & 0.0074 & 0.0074 & 0.0076 & 0.0081
        \\
        \midrule
        PATROL~\cite{ding2024patrol} &79.18& 0.0099& 0.0118 & 0.0114 & 0.0137 \\
        PATROL+CEM& 79.88 & 0.0166 & 0.0185 &0.0184 &0.0192 
        \\
        \midrule
        ResSFL~\cite{li2022ressfl} &79.60& 0.0094 & 0.0112 & 0.0111 &0.0142\\
        ResSFL+CEM&79.54 & 0.0128& 0.0148 &0.0143&0.0161  \\
        \midrule
        Noise\_Nopeek~\cite{titcombe2021practical}
        & 82.06 & 0.0052 & 0.0052 & 0.0053 & 0.0056\\
        Noise\_Nopeek+CEM& 81.96 & 0.0076 & 0.0075 & 0.0078 &0.0090\\
        \midrule
        Noise\_ARL~\cite{jeong2023noisy} &80.14& 0.0122 &0.0155 & 0.0132 &0.0158 \\
        Noise\_ARL+CEM& 80.33 & 0.0182 & 0.0211 &0.0212 &0.0231\\
        \midrule\rowcolor{gray!20}
        \textbf{Average w/o CEM}
        & 81.26 &0.0069 & 0.0079 & 0.0076 & 0.088\\
        \rowcolor{gray!20} \textbf{Average w/ CEM} &\textbf{81.38} & \textbf{0.0101} &\textbf{0.0111} & \textbf{0.0114} &\textbf{0.0123}\\
        \bottomrule   
    \end{tabular}}
    \vspace{-5mm}
    \label{tab:3}
\end{table}
\vspace{0.5mm}
\noindent \textbf{Results on TinyImageNet:}
The performance of different defense methods on the TinyImageNet dataset is presented in \tableautorefname~\ref{tab:2}, where we provide a comparative analysis of each method, both before and after integrating the proposed CEM algorithm.
The results indicate again that the integration of the CEM algorithm brings substantial inversion robustness gain.
Integrating the CEM algorithm \textbf{yields an average increase in the reconstruction MSE of 19.4\% for training data and 17.7\% for inference data}, without compromising prediction accuracy.
Meanwhile, the combination of ARL-based methods such as PATROL and Noise\_ARL with CEM achieves the best utility and robustness trade-off bringing a great robustness gain with only a marginal accuracy drop.

\begin{figure*}[t]
    \centering
    \begin{subfigure}[b]{0.32\linewidth}
        \centering
        \includegraphics[width=\linewidth]{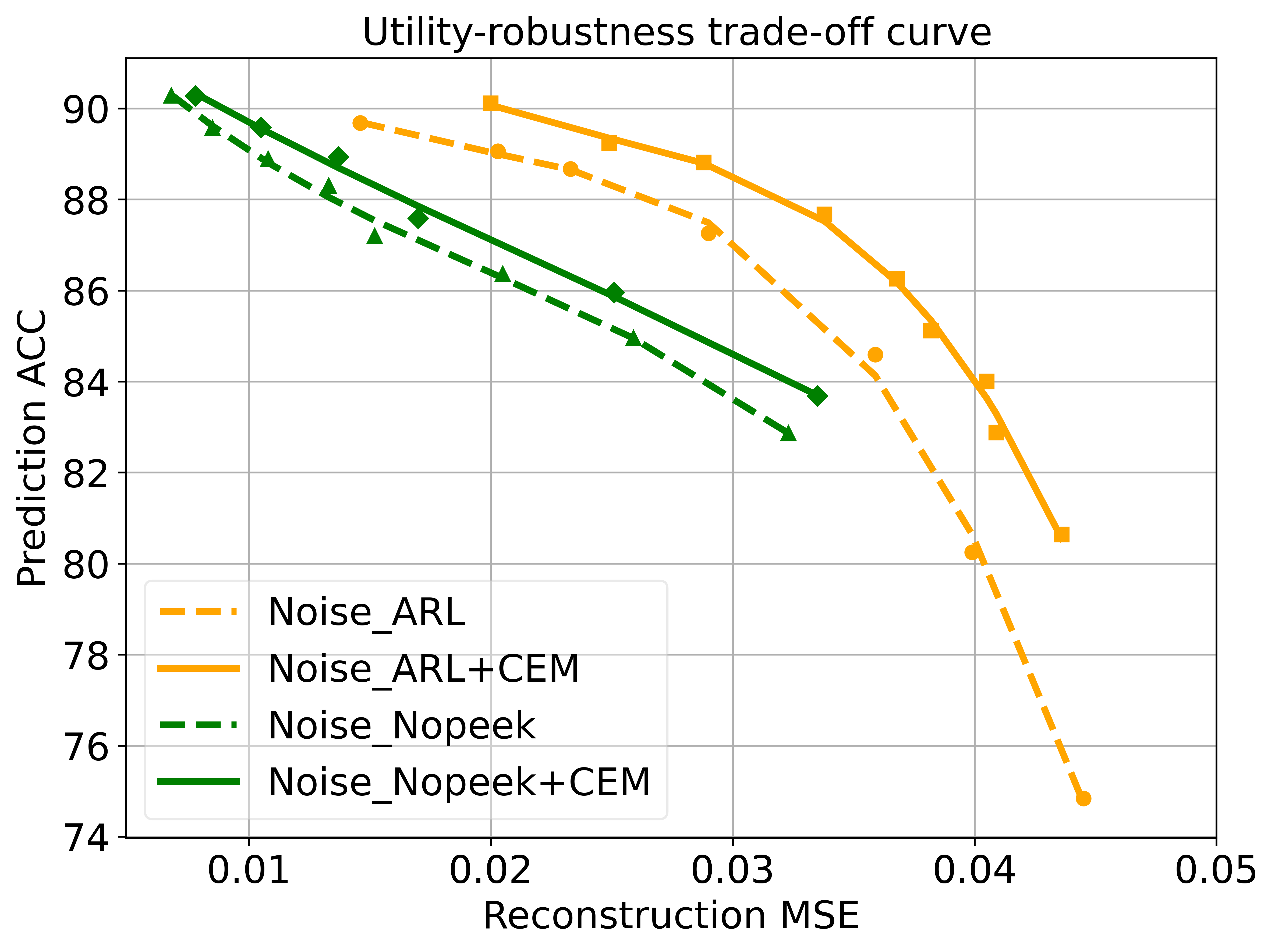}
        \vspace{-4mm}
        \caption{Noise variance varying from 0.01 to 0.5}
        \label{fig:hyper a}
    \end{subfigure}
    \hfill
    \begin{subfigure}[b]{0.32\linewidth}
        \centering
        \includegraphics[width=\linewidth]{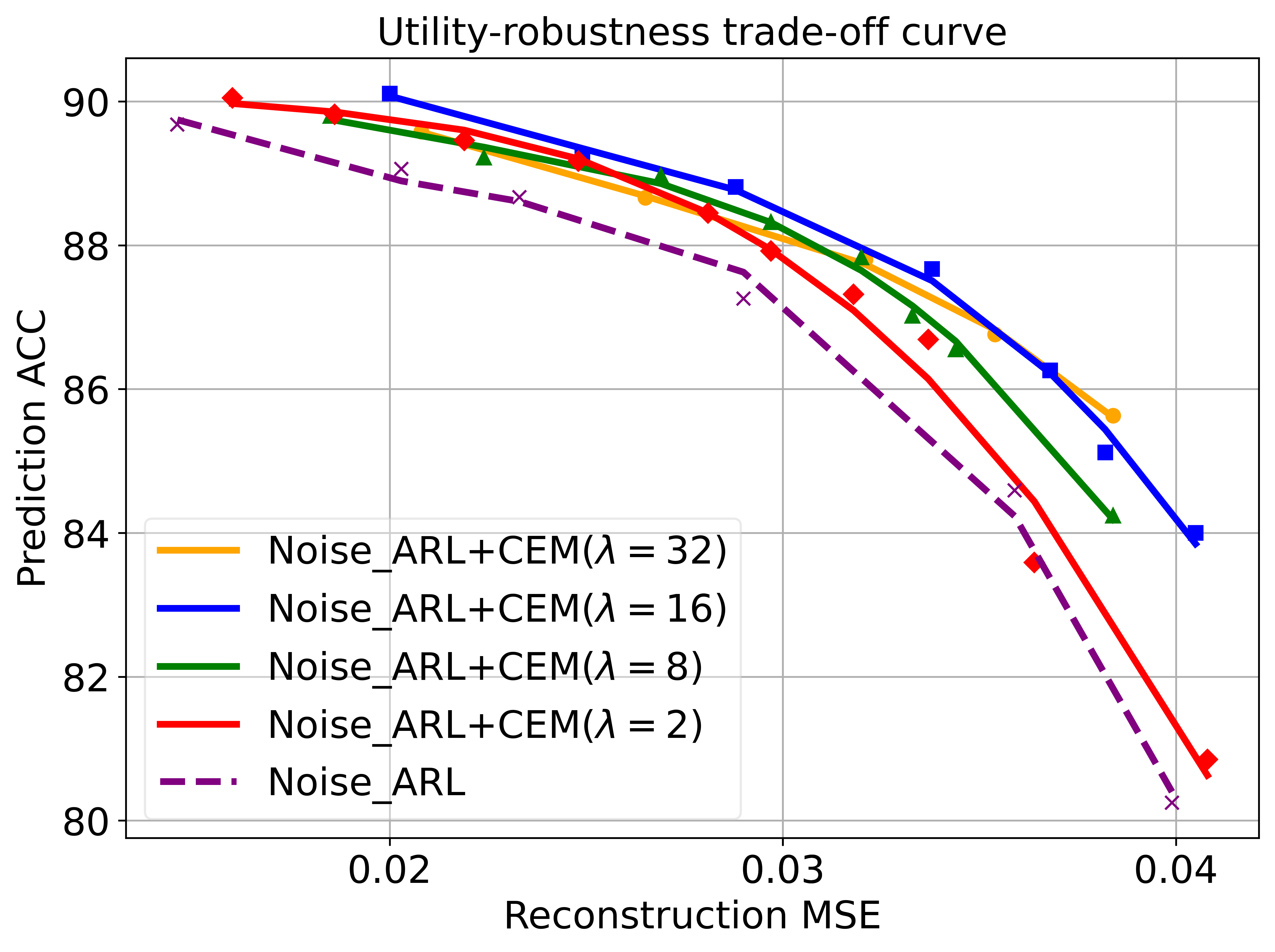}
        \vspace{-4mm}
        \caption{comparing results of different value of $\lambda$}
        \label{fig:hyper b}
    \end{subfigure}
    \hfill
    \begin{subfigure}[b]{0.32\linewidth}
        \centering
        \includegraphics[width=\linewidth]{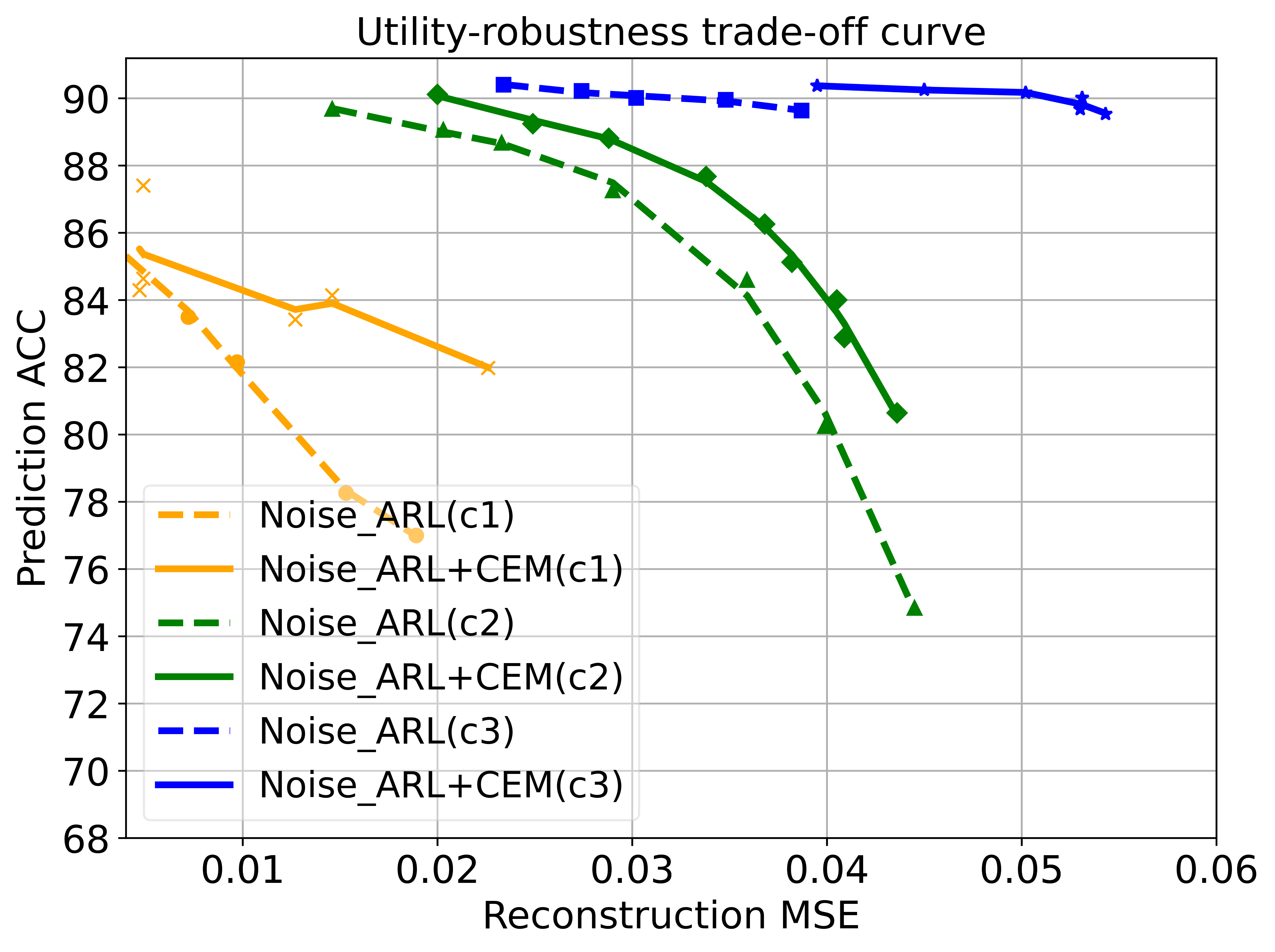}
        \vspace{-4mm}
        \caption{comparing results of different partitions}
        \label{fig:hyper c}
    \end{subfigure}
    \vspace{-3mm}
    \caption{The effect of different hyperparameters on the performance of the proposed CEM algorithm. The effect of noise strength, $\lambda$, and partitioning schemes are demonstrated by the utility-robustness trade-off curve.}
    \label{fig:three_subfigures_subcaption}
    \vspace{-5mm}
\end{figure*}

\noindent \textbf{Results on Facescrub:}
The performance of different defense methods on the Facescrub dataset is presented in \tableautorefname~\ref{tab:3}.
Unlike their performance on general object classification datasets, models in this task demonstrate increased vulnerability to MIAs.
The evaluated defense methods exhibit a notable accuracy reduction as a trade-off for increased inversion robustness.
This may caused by the strong correlation between facial features and identity information.
In this dataset, integrating the CEM algorithm with existing defense methods brings a more pronounced improvement in inversion robustness. 
Specifically, the incorporation of the CEM algorithm \textbf{yields an average increase in inversion robustness of 48.2\% for training data and 40.1\% for inference data} while maintaining prediction accuracy.
Meanwhile, the combination of ARL-based methods also achieves the best utility and robustness trade-off, which leads to a great robustness gain with only a marginal accuracy drop.  

\noindent \textbf{Results on CIFAR100:} The results on the CIFAR-100 dataset are provided in the supplementary materials S.5.

\begin{figure}[t]
    \centering
\includegraphics[width=0.95\linewidth]{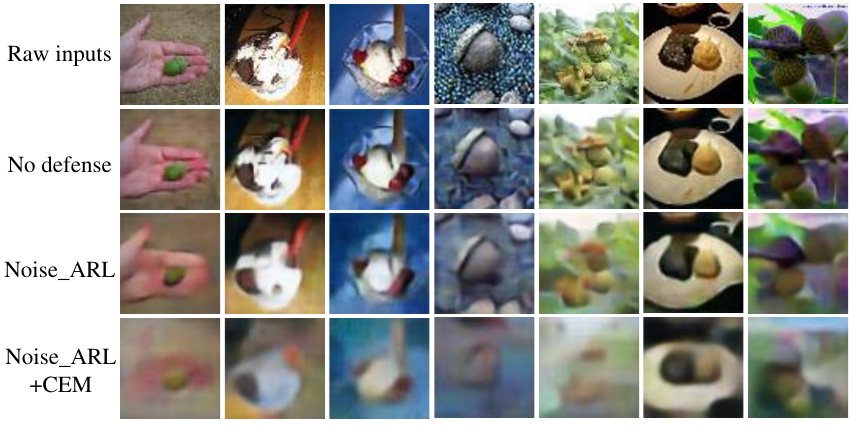} 
    \vspace{-4mm}
    \caption{The visualized result. The last three rows are inputs reconstructed by MIA.}
    \label{fig:5 visualized}
    \vspace{-7mm}
\end{figure}

\noindent \textbf{Visualized results:} We present the visualized MIA reconstruction on the TinyImageNet dataset in \figureautorefname~\ref{fig:5 visualized}. The results illustrate that incorporating the proposed CEM algorithm provides enhanced user input protection, resulting in decreased inverse reconstruction fidelity.

\vspace{-1mm}
\subsection{Ablation Study: Effect of Hyperparameters}
\vspace{-2mm}
To rigorously evaluate the impact of different hyperparameters, such as the noise strength and CEM weight factor $\lambda$, on the performance of the CEM algorithm, we conduct experiments under various settings of hyperparameters to demonstrate its robustness, versatility, and adaptability in diverse scenarios.

\noindent\textbf{The effect of noise strength:} As discussed in Subsection~\ref{trade-off}, the intensity of additive Gaussian noise is pivotal in balancing a good trade-off between utility and robustness.
We thereby evaluate the performance of the CEM algorithm under various noise strengths.
We utilize the Noise\_Nopeek~\cite{titcombe2021practical} and Noise\_ARL~\cite{jeong2023noisy} as the basic defense strategies $\mathcal{M}$.
In \figureautorefname~\ref{fig:hyper a}, we present the prediction accuracy Vs. reconstruction MSE curves on the CIFAR-10 dataset by varying the noise variance from 0.001 to 0.5.
Multiple models were trained to analyze the impact comprehensively.
The results demonstrate that the noise variance significantly influences the trade-off between feature utility and robustness.
The proposed CEM algorithm exhibits robust performance across the whole range of noise intensities.
Integrating the CEM algorithm consistently achieves a more favorable balance between utility and robustness.

\noindent\textbf{The effect of $\lambda$:}
In \equationautorefname~\ref{eq:joint_loss}, the weight factor $\lambda$ is important in adjusting the conditional entropy maximization during the joint optimization. 
To investigate the influence of $\lambda$, we use the Noise\_ARL method~\cite{jeong2023noisy}, which achieves a comparatively high performance as our baseline strategy $\mathcal{M}$.
We then evaluate its performance when integrated with the CEM algorithm, using $\lambda$ values of 0, 2, 8, 16, and 32.
We present the prediction accuracy and reconstruction MSE curve in \figureautorefname~\ref{fig:hyper b}, by varying the noise variance from 0.001 to 0.5.
The results demonstrate that increasing 
$\lambda$ from 0 to 16 markedly enhances the robustness and utility trade-off.

\noindent\textbf{Different partitioning mechanisms:} Encoders with more layers extract more task-specific features from input data, thereby providing improved robustness against MIAs.
To illustrate the effectiveness of the proposed CEM under different levels of redundancy, we evaluate its performance across multiple partitioning schemes.
Specifically, we use the VGG11 architecture as the backbone, partitioned at the 1st, 2nd, and 3rd convolutional layers (denoted as $c_1$, $c_2$, and $c_3$), and assess the performance of CEM in combination with Noise\_ARL~\cite{jeong2023noisy}.
We present the prediction accuracy Vs. reconstruction MSE curve in \figureautorefname~\ref{fig:hyper c} using the same strategy.
The results demonstrate a significant performance advantage with deeper partitioning: specifically, partitioning at the third convolutional layer achieves substantial inversion robustness while maintaining prediction accuracy without degradation.
Furthermore, the CEM algorithm exhibits high adaptability, consistently enhancing the robustness-utility trade-off across different partitioning schemes.

\noindent \textbf{Analysis of the efficiency:} The computational complexity on the local encoder and the cloud server across various methods is presented in the supplementary material S.6. The results confirm that integrating the proposed CEM does not introduce any additional computational overhead or latency.
\vspace{-2mm}
\section{Conclusion}
\vspace{-1mm}
This work addresses problems of inversion robustness in collaborative inference systems, where MIAs can exploit subtle signals within these intermediate features to reconstruct high-fidelity inputs.
Existing obfuscation defenses seek to protect privacy by empirically eliminating feature redundancy. 
However, the precise quantification of such redundancy is lacking, and rigorous mathematical analysis is needed to elucidate the relation between redundancy minimization and enhanced inversion robustness.
To solve that, we prove that the conditional entropy of inputs given intermediate features provides a lower bound on the reconstruction MSE.
Based on that, we derive a differentiable lower bound on this conditional entropy using Gaussian mixture estimation, making it amenable for efficient optimization through backpropagation.
Building on this, we propose a versatile conditional entropy maximization (CEM) algorithm that can be easily plugged into existing methods.
Comprehensive Experiments demonstrate that the proposed CEM significantly and consistently boosts robustness while maintaining feature utility and computational efficiency.

\section*{Acknowledgement}
This work was carried out at the Rapid-Rich Object Search (ROSE) Lab, School of Electrical \& Electronic Engineering, Nanyang Technological University (NTU), Singapore. This research is supported by the National Research Foundation, Singapore and Infocomm Media Development Authority under its Trust Tech Funding Initiative, the Basic and Frontier Research Project of PCL, the Major Key Project of PCL, Guangdong Basic and Applied Basic Research Foundation under Grant 2024A1515010454, the Program of Beijing Municipal Science and Technology Commission Foundation (No.Z241100003524010),  and in part by the PKU-NTU Joint Research Institute (JRI) sponsored by a donation from the Ng Teng Fong Charitable Foundation. Any opinions, findings and conclusions or recommendations expressed in this material are those of the author(s) and do not reflect the views of National Research Foundation, Singapore and Infocomm Media Development Authority.

{
    \small
    \bibliographystyle{ieeenat_fullname}
    \bibliography{main}
}
\newpage
\clearpage
\setcounter{page}{1}
\maketitlesupplementary

\section{Proof of Theorem 1}
Given the covariance matrix $Cov(\vx|\vz)$, the conditional entropy $\mathcal{H}(\vx|\vz)$ satisfies that:
\begin{equation}
\mathcal{H}(\vx|\vz)\le \mathbb{E}_{\mathcal{Z}}\left [\frac{1}{2} \log\left ( (2\pi{e})^d\text{det}\left ( Cov(\vx|\vz)\right )  \right )\right ],
\label{eq:entropy upper bound}
\end{equation}
where $\text{det}$ denotes the determinant of the matrix.  
\equationautorefname~\ref{eq:entropy upper bound} formalizes the principle that the Gaussian distribution achieves the maximum entropy among all distributions with a given covariance.

\noindent Let $\lambda \in  \left \{\lambda_1,...,\lambda_d  \right \} $ denote the eigenvalues of the matrix $Cov(\vx|\vz)$. 
It follows that: 
\begin{equation}
    \text{det}(Cov(\vx|\vz))=\prod_{i=1}^{d}\lambda_i, \: Tr(Cov(\vx|\vz))=\sum_{i=1}^{d}\lambda_i .
    \label{eq:det and trace}
\end{equation}
Using the Jensen inequality, we can get:
\begin{equation}
\mathcal{H}(\vx|\vz)\le \mathbb{E}_{\mathcal{Z}}\left [\frac{d}{2} \log\left ( 2\pi{e}\frac{{Tr}\left ( Cov(\vx|\vz)\right )}{d}\right ) \right ],
\label{eq:entropy upper bound2}
\end{equation}
As the $\log$ function is concave, we can get:
\begin{align}
\mathcal{H}(\vx|\vz)&\le \frac{d}{2} \log\left ( \frac{2\pi{e}}{d}\mathbb{E}_{\mathcal{Z}}\left [  {Tr}\left ( Cov(\vx|\vz)\right ) \right]\right ),
\\
&\le \frac{d}{2} \log\left ( 2\pi{e}\varepsilon \right ),
\label{eq:entropy upper bound3}
\end{align}
which concludes the proof of Theorem~1.

\section{Proof of Theorem 2}

Proposition 2 illustrates that maximizing the $H(\vx|\vz)$ is equivalent to minimizing the mutual information $\mathcal{I}(\vz;{\hat{\vz}})$, which is:

\begin{equation}
\mathcal{I}(\vz;\hat{\vz})=\mathcal{H}({\vz})-\mathcal{H}({\vz}|\hat{\vz})
=\mathcal{H}({\vz})-\frac{1}{2}\log((2\pi e)^d \left| \Sigma_p\right|).
    \label{eq:opt_fea_rob}
\end{equation}

\noindent It is hard to give a closed-form representation of $\mathcal{H}(\vz)$ when $\vz$ follows the Gaussian mixture distribution. In~\cite{huber2008entropy}, an upper bound of $\mathcal{H}(\vz)$ is given by:

\begin{equation}
\mathcal{H}(\vz) \le \sum_{i=1}^{k} \pi_i\left (-\log(\pi_i) + \frac{1}{2}\log((2\pi e)^d \left| \Sigma_i+\Sigma_p\right|  )\right). 
    \label{eq:GMM_upperbound}
\end{equation}

\noindent Therefore, we claim that the mutual information $\mathcal{I}(\vz;{\hat{\vz}})$ satisfies that:
\begin{equation}
\mathcal{I}(\vz;{\hat{\vz}}) \le \sum_{i=1}^{k} \pi_i\left (-\log(\pi_i) + \frac{1}{2}\log(\frac{ \left| \Sigma_i+\Sigma_p\right|}{\left|\Sigma_p\right|}  )\right), 
    \label{eq:MI_upperbound_prove}
\end{equation}
which concludes the proof of Theorem 2.

\section{Architecture and Training details of Inversion Models}
\textbf{Decoding-based inversion model:} We allow the inversion adversary to utilize a complex inversion model to reconstruct the original inputs. Specifically, we utilize a decoder network with 8 concatenated residual blocks and the corresponding number of transpose convolutional blocks to recover the original size of the input. Each convolutional layer has 64 channels. 
\begin{figure}[h]
    \centering
    \includegraphics[width=1\linewidth]{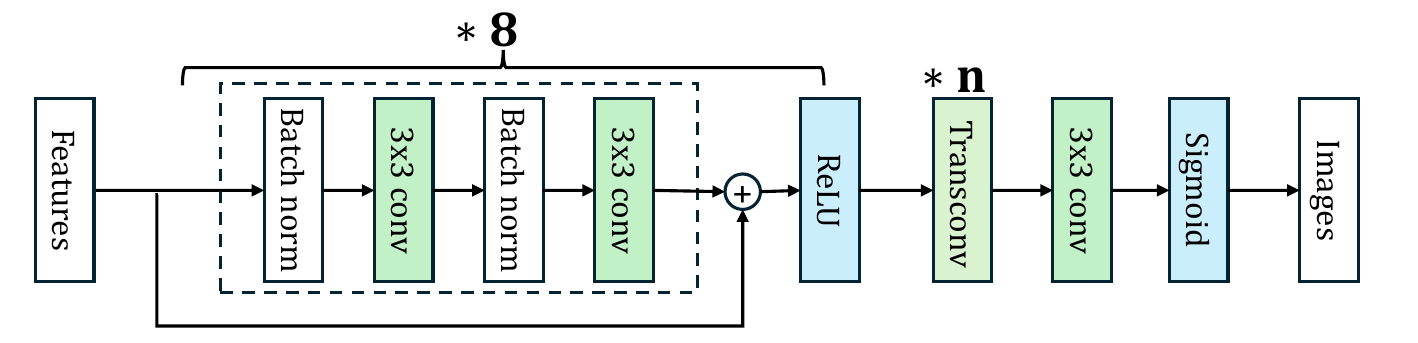}
    \caption{The structure of the decoding network.}
    \label{fig:dec_net}
\end{figure}

\noindent The architecture of the decoding-based inversion model is illustrated in \figureautorefname~\ref{fig:dec_net}. Specifically, the initial eight residual blocks are designed to process the extracted features, while the transposed convolutional layers progressively upsample the feature maps to match the dimensions of the original image. The final convolutional layer performs the concluding processing, and the application of the Sigmoid activation function normalizes the output to the range [0, 1]. The decoder comprises approximately 711.54k trainable parameters and requires 90.66 MMAC operations for computation. We train the decoder model for 50 epochs using the Adam optimizer with an initial learning rate of 0.005. 

\vspace{1mm}
\noindent \textbf{GAN-based inversion model:} We follows the methodology outlined in~\cite{zhang2020secret} to implement the GAN-based inversion attack. 
Concurrently, we replace the original generator with the proposed decoding-based network, which is architecturally more complex and delivers superior performance.

\vspace{1mm}
\noindent The GAN inversion model is trained for 150 epochs using the Adam optimizer with an initial learning rate of 0.005.
Additionally, a MSE loss term is incorporated to regularize the training process, where we find it effectively facilitates the GAN inversion model in achieving a lower reconstruction MSE.

\section{Experimental results of SSIM and PSNR}
The experimental results of reconstruction SSIM and PSNR on the validation set on the CIFAR10, CIFAR100, TinyImageNet, and FaceScrub datasets are presented in \tableautorefname~\ref{tab:1} to~\ref{tab:4}.
We provide a comparative analysis of each method, both before and after integrating the proposed CEM algorithm.

\vspace{1mm}
\noindent The results show that the integration of the CEM algorithm consistently enhances all defense methods on four datasets.
On the CIFAR10 dataset, plugging in our proposed CEM algorithm improves the average of \textbf{SSIM from 0.673 to 0.639 and PSNR from 18.28 to 17.51}.
On the CIFAR100 dataset, plugging in our proposed CEM algorithm improves the average of \textbf{SSIM from 0.814 to 0.755 and PSNR from 23.06 to 20.63}.
On the TinyImageNet dataset, plugging in our proposed CEM algorithm improves the average of \textbf{SSIM from 0.567 to 0.523 and PSNR from 18.09 to 17.37}.
On the FaceScrub dataset, plugging in our proposed CEM algorithm improves the average of \textbf{SSIM from 0.794 to 0.752 and PSNR from 21.59 to 20.07}.

\begin{table}[h]
    \centering
    \caption{Comparative results on CIFAR10 dataset: we present the accuracy and reconstruction SSIM and PSNR on the validation set. }
    \resizebox{0.48\textwidth}{!}{
    \begin{tabular}{lccccc}
        \toprule
        \multirow{2}{*}{\textbf{Methods}} &\multirow{2}{*}{\textbf{Acc.$\uparrow$}}& \multicolumn{2}{c}{\textbf{Dec.-based MIA~\cite{li2022ressfl}}} & \multicolumn{2}{c}{\textbf{GAN-based MIA~\cite{zhang2020secret}}}
        \\
        \cmidrule(lr){3-4} \cmidrule(lr){5-6} 
        &  & \textbf{SSIM$\downarrow$} &\textbf{PSNR$\downarrow$} & \textbf{SSIM$\downarrow$} &\textbf{PSNR$\downarrow$}  \\
        \midrule
        {Bottleneck} 
        & 90.87 & 0.863 & 23.82 & 0.861 & 23.46\\
        Bottleneck+CEM& 90.69 & 0.813 & 22.36 & 0.783 &20.96\\
        \midrule
        DistCorr~\cite{vepakomma2020nopeek} &89.52& 0.779 & 20.86 & 0.780 & 20.55 \\
        DistCorr+CEM & 89.80 & 0.757 & 20.31 &0.760 &20.17 \\
        \midrule
        Dropout~\cite{he2020attacking} 
        & 87.75 & 0.729 & 19.82 & 0.733  & 19.54\\
        Dropout+CEM& 87.53 & 0.682 & 18.72 & 0.687 & 18.47\\
        \midrule
        PATROL~\cite{ding2024patrol} &89.58& 0.537& 15.33 & 0.558 & 15.12 \\
        PATROL+CEM& 89.67 & 0.506 & 14.74 &0.520 &14.59 \\
        \midrule
        ResSFL~\cite{li2022ressfl} &89.68& 0.595 & 16.19 & 0.639 &16.14\\
        ResSFL+CEM&90.11 & 0.571& 16.00 &0.583&15.63  \\
        \midrule
        Noise\_Nopeek~\cite{titcombe2021practical}
        & 87.19 & 0.664 & 17.98 & 0.668 & 17.82
        \\
        Noise\_Nopeek+CEM& 87.08 & 0.643 & 17.59 & 0.651 &17.49 \\
        \midrule
        Noise\_ARL~\cite{jeong2023noisy} &87.78& 0.501 &14.73 & 0.518 &14.65 \\
        Noise\_ARL+CEM& 87.62 & 0.484 & 14.48 &0.502 &14.40\\
        \midrule\rowcolor{gray!20}
        \textbf{Average w/o CEM}
        & 88.91 &0.667 & 18.39 & 0.680 & 18.18 \\
        \rowcolor{gray!20} \textbf{Average w/ CEM} &\textbf{88.92} & \textbf{0.637} &\textbf{17.74} & \textbf{0.641} &\textbf{17.38} \\
        \bottomrule   
    \end{tabular}}
    \label{tab:1}
\end{table}

\begin{table}[h]
    \centering
    \caption{Comparative results on CIFAR100 dataset: we present the accuracy and reconstruction SSIM and PSNR on the validation set. }
    \resizebox{0.48\textwidth}{!}{
    \begin{tabular}{lccccc}
        \toprule
        \multirow{2}{*}{\textbf{Methods}} &\multirow{2}{*}{\textbf{Acc.$\uparrow$}}& \multicolumn{2}{c}{\textbf{Dec.-based MIA~\cite{li2022ressfl}}} & \multicolumn{2}{c}{\textbf{GAN-based MIA~\cite{zhang2020secret}}}
        \\
        \cmidrule(lr){3-4} \cmidrule(lr){5-6} 
        &  & \textbf{SSIM$\downarrow$} &\textbf{PSNR$\downarrow$} & \textbf{SSIM$\downarrow$} &\textbf{PSNR$\downarrow$}  \\
        \midrule
        {Bottleneck} 
        & 68.43 & 0.975 & 31.54 & 0.970 & 30.96\\
        Bottleneck+CEM& 68.42 & 0.856 & 22.36 & 0.937 &26.98\\
        \midrule
        DistCorr~\cite{vepakomma2020nopeek} &66.21& 0.880 & 22.67 & 0.928 & 26.02 \\
        DistCorr+CEM & 66.27 & 0.812 & 21.30 &0.877 &23.46 \\
        \midrule
        Dropout~\cite{he2020attacking} 
        & 65.85 & 0.865 & 23.76 & 0.936  & 27.44\\
        Dropout+CEM& 65.92 & 0.816 & 22.21 & 0.890 & 24.94\\
        \midrule
        PATROL~\cite{ding2024patrol} &65.10& 0.478& 14.74 & 0.634 & 16.23 \\
        PATROL+CEM& 65.07 & 0.440 & 13.77 &0.603 &15.96 \\
        \midrule
        ResSFL~\cite{li2022ressfl} &66.94& 0.866 & 22.92 & 0.935 &26.98\\
        ResSFL+CEM&66.96 & 0.770& 20.31 &0.866&23.37  \\
        \midrule
        Noise\_Nopeek~\cite{titcombe2021practical}
        & 65.55 & 0.841 & 22.00 & 0.890 & 24.20
        \\
        Noise\_Nopeek+CEM& 65.33 & 0.797 & 20.80 & 0.858 &22.83 \\
        \midrule
        Noise\_ARL~\cite{jeong2023noisy} &62.58& 0.521 &15.57 & 0.691 &17.85 \\
        Noise\_ARL+CEM& 62.34 & 0.457 & 16.27 &0.599 &14.40\\
        \midrule\rowcolor{gray!20}
        \textbf{Average w/o CEM}
        & 65.81 &0.775 & 21.88 & 0.854 & 24.24 \\
        \rowcolor{gray!20} \textbf{Average w/ CEM} &\textbf{65.75} & \textbf{0.706} &\textbf{19.57} & \textbf{0.804} &\textbf{21.70} \\
        \bottomrule   
    \end{tabular}}
    \label{tab:2}
\end{table}

\begin{table}[h]
    \centering
    \caption{Comparative results on TinyImageNet dataset: we present the accuracy and reconstruction SSIM and PSNR on the validation set. }
    \resizebox{0.48\textwidth}{!}{
    \begin{tabular}{lccccc}
        \toprule
        \multirow{2}{*}{\textbf{Methods}} &\multirow{2}{*}{\textbf{Acc.$\uparrow$}}& \multicolumn{2}{c}{\textbf{Dec.-based MIA~\cite{li2022ressfl}}} & \multicolumn{2}{c}{\textbf{GAN-based MIA~\cite{zhang2020secret}}}
        \\
        \cmidrule(lr){3-4} \cmidrule(lr){5-6} 
        &  & \textbf{SSIM$\downarrow$} &\textbf{PSNR$\downarrow$} & \textbf{SSIM$\downarrow$} &\textbf{PSNR$\downarrow$}  \\
        \midrule
        {Bottleneck} 
        & 52.77 & 0.666 & 19.87 & 0.698 & 20.36\\
        Bottleneck+CEM& 52.52 & 0.593 & 18.86 & 0.623 &19.03\\
        \midrule
        DistCorr~\cite{vepakomma2020nopeek} &51.79& 0.598 & 18.38 & 0.627 & 18.66 \\
        DistCorr+CEM & 51.78 & 0.527 & 17.21 &0.553 &17.74 \\
        \midrule
        Dropout~\cite{he2020attacking} 
        & 50.72 & 0.548 & 17.87 & 0.567  & 18.21\\
        Dropout+CEM& 50.75 & 0.511 & 17.30 & 0.533 & 17.64\\
        \midrule
        PATROL~\cite{ding2024patrol} & 51.75& 0.512& 17.28 & 0.536 & 17.54 \\
        PATROL+CEM& 51.64 & 0.489 & 16.79 &0.524 &17.23 \\
        \midrule
        ResSFL~\cite{li2022ressfl} & 51.99& 0.522 & 17.64 &0.552 &17.93\\
        ResSFL+CEM& 52.07 & 0.495& 17.12 &0.521&17.61  \\

        \midrule    Noise\_Nopeek~\cite{titcombe2021practical}&51.63& 0.578 & 18.04 & 0.588 &18.18\\
        Noise\_Nopeek+CEM&52.01 & 0.527& 17.44 &0.551&17.77  \\
        \midrule
        Noise\_ARL~\cite{jeong2023noisy} & 51.03& 0.463 &16.49 & 0.486 &16.88 \\
        Noise\_ARL+CEM&  50.85 & 0.428 & 15.67 &0.441 &15.96\\
        \midrule\rowcolor{gray!20}
        \textbf{Average w/o CEM}
        & \textbf{51.66} &0.555 & 17.93 & 0.579 & 18.25 \\
        \rowcolor{gray!20} \textbf{Average w/ CEM} &51.62 & \textbf{0.510} &\textbf{17.19} & \textbf{0.535} &\textbf{17.56} \\
        \bottomrule   
    \end{tabular}}
    \label{tab:3}
\end{table}

\begin{table}[h]
    \centering
    \caption{Comparative results on FaceScrub dataset: we present the prediction accuracy and reconstruction SSIM and PSNR on the validation set. }
    \resizebox{0.48\textwidth}{!}{
    \begin{tabular}{lccccc}
        \toprule
        \multirow{2}{*}{\textbf{Methods}} &\multirow{2}{*}{\textbf{Acc.$\uparrow$}}& \multicolumn{2}{c}{\textbf{Dec.-based MIA~\cite{li2022ressfl}}} & \multicolumn{2}{c}{\textbf{GAN-based MIA~\cite{zhang2020secret}}}
        \\
        \cmidrule(lr){3-4} \cmidrule(lr){5-6} 
        &  & \textbf{SSIM$\downarrow$} &\textbf{PSNR$\downarrow$} & \textbf{SSIM$\downarrow$} &\textbf{PSNR$\downarrow$}  \\
        \midrule
        {Bottleneck} 
        & 85.12 & 0.898 &26.02 & 0.864 & 25.68\\
        Bottleneck+CEM& 85.00 & 0.860 & 24.45 & 0.821&24.31\\
        \midrule
        DistCorr~\cite{vepakomma2020nopeek} &83.42& 0.853 & 24.20 & 0.848 & 23.87 \\
        DistCorr+CEM & 83.78 & 0.833 & 23.27 &0.795 &21.30 \\
        \midrule
        Dropout~\cite{he2020attacking} 
        & 79.30 & 0.813 & 22.83 & 0.808 & 22.44\\
        Dropout+CEM& 79.19 & 0.776 & 21.30 & 0.771 & 20.91\\
        \midrule
        PATROL~\cite{ding2024patrol} &79.18& 0.737& 19.28 & 0.731 & 18.63 \\
        PATROL+CEM& 79.88 & 0.685 & 17.32 &0.681 &17.16 \\
        \midrule
        ResSFL~\cite{li2022ressfl} &79.60& 0.745 & 19.50 & 0.736 &18.47\\
        ResSFL+CEM&79.54 & 0.713& 18.29 &0.710& 17.93  \\
        \midrule
        Noise\_Nopeek~\cite{titcombe2021practical}
        & 82.06 & 0.844 & 22.83 & 0.839 & 22.51
        \\
        Noise\_Nopeek+CEM& 81.96 & 0.784 & 21.24 & 0.777 &20.45 \\
        \midrule
        Noise\_ARL~\cite{jeong2023noisy} &80.14& 0.705 &18.09 & 0.710 &18.01 \\
        Noise\_ARL+CEM& 80.33 & 0.669 & 16.75 &0.660 &16.36\\
        \midrule\rowcolor{gray!20}
        \textbf{Average w/o CEM}
        & 81.26 &0.799 & 21.82 & 0.790 & 21.37 \\
        \rowcolor{gray!20} \textbf{Average w/ CEM} &\textbf{81.38} & \textbf{0.760} &\textbf{20.37} & \textbf{0.745} &\textbf{19.77} \\
        \bottomrule   
    \end{tabular}}
    \label{tab:4}
\end{table}
\section{Experimental results on CIFAR100}
The performance of different defense methods on the CIFAR100 dataset is presented in \tableautorefname~\ref{tab:5}, where we provide a comparative analysis of each method, both before and after integrating the proposed CEM algorithm.
The results indicate again that the integration of the CEM algorithm brings substantial inversion robustness gain.
Integrating the CEM algorithm \textbf{yields an average increase in the reconstruction MSE of 40.5\% for training data and 44.8\% for inference data}, without compromising prediction accuracy.

\begin{table}[h]
    \centering
    \caption{Comparative results on CIFAR100 dataset: we present the accuracy and reconstruction MSE of different defense methods using the VGG11 model with and without integrating the proposed CEM. }
    \resizebox{0.48\textwidth}{!}{
    \begin{tabular}{lcccccc}
        \toprule
        \multirow{2}{*}{\textbf{Methods}} &\multirow{2}{*}{\textbf{Acc.$\uparrow$}}& \multicolumn{2}{c}{\textbf{Dec.-based MSE~\cite{li2022ressfl}}} & \multicolumn{2}{c}{\textbf{GAN-based MSE~\cite{zhang2020secret}}}
        \\
        \cmidrule(lr){3-4} \cmidrule(lr){5-6} 
        &  & \textbf{Train$\uparrow$} &\textbf{Infer$\uparrow$} & \textbf{Train$\uparrow$} &\textbf{Infer$\uparrow$}\\
        \midrule
        {Bottleneck} 
        & 68.43 & 0.0007 & 0.0007 & 0.0009 & 0.0008\\
        Bottleneck+CEM& 68.42 & 0.0058 & 0.0059 & 0.0019 &0.0020
        \\
        \midrule
        DistCorr~\cite{vepakomma2020nopeek} &66.21& 0.0054 & 0.0055 & 0.0023 & 0.0025 \\
        DistCorr+CEM & 66.27 & 0.0074 & 0.0075 &0.0044 &0.0045 \\
        \midrule
        Dropout~\cite{he2020attacking} 
        & 65.85 & 0.0042 & 0.0043 & 0.0016 & 0.0018\\
        Dropout+CEM& 65.92 & 0.0060 & 0.0061 & 0.0031 & 0.0032	
        \\
        \midrule
        PATROL~\cite{ding2024patrol} &65.10& 0.0335& 0.0346 & 0.0196 & 0.0238 \\
        PATROL+CEM& 65.07 & 0.0419&  0.0423&0.0235 &0.0253 
        \\
        \midrule
        ResSFL~\cite{li2022ressfl} &66.94& 0.0051 & 0.0052 & 0.0019 &0.0020\\
        ResSFL+CEM&66.96 & 0.0093& 0.0095 &0.0043&0.0046  \\
        \midrule
        Noise\_Nopeek~\cite{titcombe2021practical}
        & 65.55 & 0.0063 & 0.0063 & 0.0037 & 0.0038\\
        Noise\_Nopeek+CEM& 65.33 & 0.0083 & 0.0082 & 0.0052 &0.0052\\
        \midrule
        Noise\_ARL~\cite{jeong2023noisy} &62.58& 0.0270 &0.0277 & 0.0143 &0.0164 \\
        Noise\_ARL+CEM& 62.34 & 0.0373 &0.0380 &0.0219 &0.0236\\
        \midrule\rowcolor{gray!20}
        \textbf{Average w/o CEM}
        & \textbf{65.81} &0.0117 & 0.0120 & 0.0063 & 0.0073\\
        \rowcolor{gray!20} \textbf{Average w/ CEM} &65.75 & \textbf{0.0165} &\textbf{0.0168} & \textbf{0.0095} &\textbf{0.0102}\\
        \bottomrule   
    \end{tabular}}
    \label{tab:5}
\end{table}

\section{Analysis of the efficiency}
We evaluate the inference time and model size of the local encoder and cloud server using one RTX 4090 GPU, with detailed results presented in \tableautorefname~\ref{tab:6}.
Notably, all methods, except for PATROL, exhibit identical inference efficiency and parameter counts. 
The inference time is measured by processing a batch of 128 images over 100 times

\begin{table}[h]
    \centering
    \caption{Analysis of the efficiency. }
    \resizebox{0.48\textwidth}{!}{
    \begin{tabular}{ccccccc}
    \toprule
    \multirow{2}{*}{Method}&\multicolumn{3}{c}{\textbf{Local encoder}}&\multicolumn{3}{c}{\textbf{Could server}}
    \\
    \cmidrule(lr){2-4} \cmidrule(lr){5-7} 
    &  \textbf{Parameters} &\textbf{Flops} & \textbf{Infere time} &\textbf{Parameters} &\textbf{Flops} & \textbf{Infere time}\\
    \midrule
    PATROL & 0.083M & 27.3 MAC & 65.61 ms & 9.78M & 136.38MAC & 175.92ms
    \\
    \midrule
    OTHERS & 0.085M &21.9 MAC & 55.16 ms& 9.78M & 133.59MAC & 169.87ms
    \\
    \bottomrule
    \end{tabular}}
    \label{tab:6}
\end{table}

\end{document}